\documentclass[10pt,twocolumn,letterpaper]{article}

\usepackage{iccv}
\usepackage{epsfig}
\usepackage{graphicx}
\usepackage{times}
\usepackage{url}
\usepackage[utf8]{inputenc} 
\usepackage[T1]{fontenc}    
\usepackage{booktabs}       
\usepackage{amsfonts}       
\usepackage{amsmath}
\usepackage{amssymb}
\usepackage{amsthm}
\usepackage{nicefrac}       
\usepackage{microtype}      
\usepackage{dsfont}
\usepackage{multirow}
\usepackage{xcolor}
\usepackage[export]{adjustbox}
\usepackage{diagbox}
\usepackage{bm}
\usepackage{algorithm}
\usepackage{algorithmic}
\usepackage{tabularx}
\usepackage{wrapfig}
\usepackage{array, boldline, makecell}
\usepackage{color, colortbl}
\definecolor{Gray}{gray}{0.9}
\usepackage{pifont}
\newcommand{\cmark}{\ding{51}}%
\newcommand{\xmark}{\ding{55}}%
\definecolor{corange}{RGB}{255,189,132}
\definecolor{cyellow}{RGB}{255,224, 159}
\definecolor{cgreen}{RGB}{102, 178, 160}
\definecolor{cblue}{RGB}{92, 197, 227}
\newcommand{\sqboxs}{1.8ex}
\newcommand{\sqbox}[1]{\textcolor{#1}{\rule{\sqboxs}{\sqboxs}}}

\usepackage[pagebackref=true,breaklinks=true,letterpaper=true,colorlinks,bookmarks=false]{hyperref}

\iccvfinalcopy 

\def\iccvPaperID{****} 

\ificcvfinal\fi

\begin{document}

\title{AutoTaskFormer: Searching Vision Transformers for Multi-task Learning}

\author {
    Yang Liu\textsuperscript{\rm 1}\thanks{This work was done during an internship at Microsoft Research Asia}, Shen Yan\textsuperscript{\rm 2}, Yuge Zhang\textsuperscript{\rm 3}, Kan Ren\textsuperscript{\rm 3}, Quanlu Zhang\textsuperscript{\rm 3}, Zebin Ren\textsuperscript{\rm 4}\footnotemark[1], Deng Cai\textsuperscript{\rm 1}, Mi Zhang\textsuperscript{\rm 5}\\
    \textsuperscript{\rm 1}Zhejiang University \quad \textsuperscript{\rm 2}Michigan State University \quad \textsuperscript{\rm 3}Microsoft Research Asia \\
    \textsuperscript{\rm 4}Vrije Universiteit Amsterdam \quad \textsuperscript{\rm 5}The Ohio State University\\
    {\tt\normalsize lyng\_95@zju.edu.cn \quad Quanlu.Zhang@microsoft.com}
}

\maketitle
\ificcvfinal\thispagestyle{empty}\fi

\begin{abstract}
Vision Transformers have shown great performance in single tasks such as classification and segmentation. However, real-world problems are not isolated, which calls for vision transformers that can perform multiple tasks concurrently. Existing multi-task vision transformers are handcrafted and heavily rely on human expertise. In this work, we propose a novel one-shot neural architecture search framework, dubbed AutoTaskFormer (\textbf{Auto}mated Multi-\textbf{Task} Vision Trans\textbf{Former}), to automate this process. AutoTaskFormer not only identifies the weights to share across multiple tasks automatically, but also provides thousands of well-trained vision transformers with a wide range of parameters (\eg, number of heads and network depth) for deployment under various resource constraints. Experiments on both small-scale (2-task Cityscapes and 3-task NYUv2) and large-scale (16-task Taskonomy) datasets show that AutoTaskFormer outperforms state-of-the-art handcrafted vision transformers in multi-task learning. The entire code and models will be open-sourced.
\end{abstract}

\section{Introduction}

Vision Transformers have emerged as successful architectures in various domains, including image classification, object detection, and segmentation \cite{dosovitskiy2020image,liu2021swin,touvron2021training}. However, real-world problems are often interrelated, which necessitates the development of vision transformers capable of performing multiple tasks simultaneously. Autonomous driving is an example that involves multiple tasks including simultaneous lane detection, semantic understanding, obstacle detection, and depth estimation. Multi-task vision transformers leverage shared knowledge among tasks, allow for a reduction of storage and latency during inference, and yield superior performance compared to their single-task counterparts.

Typically, the architectures of such multi-task vision transformers have been handcrafted in the literature \cite{chen2021pre, mohamed2021spatio, seong2019video, bhattacharjee2022mult, ye2022inverted, ye2023taskprompter}. For instance, IPT \cite{chen2021pre} proposes a multi-head multi-tail architecture for four low-level tasks. MulT \cite{bhattacharjee2022mult} adopts a shared transformer encoder and six task-specific decoders for high-level dense predictions. InvPT \cite{ye2022inverted} simultaneously models spatial positions and multiple tasks in a unified framework. TaskPrompter \cite{ye2023taskprompter} designs a set of spatial-channel task
prompts and learn their spatial and channel interactions within each transformer layer.

Those handcrafted designs, unfortunately, are limited by two primary constraints. 
First, manual architecture exploration will likely exceed human design capabilities due to extremely large design spaces. For example, how to choose the best sharing patterns across various tasks and decide the optimal embedding dimensions and the number of heads within each layer. These challenges are magnified as the number of tasks increases.
Second, handcraft designs are generally designed for specific multi-task scenarios. Since weight sharing patterns are diverse for different tasks, the multi-task model manually designed for one scenario is hard to generalize to other multi-task scenarios.

\begin{figure}[t]
    \begin{center}
     \includegraphics[width=0.45\textwidth]{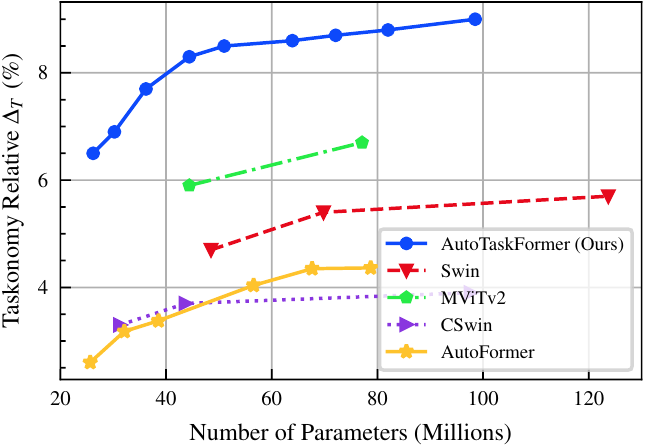}
     \end{center}
     \caption{Comparison between AutoTaskFormer and existing transformer-based and NAS-based models on 16-task Taskonomy \cite{zamir2018taskonomy}. We use the relative performance metric $\Delta_T$ with respect to the same single-task baseline to compare different models.\label{fig:pareto}}
\end{figure}

Neural architecture search (NAS) has been proposed to automate the neural architecture design to mitigate manual efforts. Recent NAS advances \cite{chen2021autoformer,cai2020once} advocate training one high-quality supernet containing thousands of high-performing subnets that can be directly searched and deployed without extra fine-tuning or re-training. It is natural to ask whether it is possible to obtain a high-quality \textit{multi-task supernet} where every \textit{multi-task subnet} exhibits decent performance. When investigating this problem, we face two major hurdles. First, designing a NAS search space that allows multiple tasks to operate simultaneously while preserving each task's performance. Second, establishing a process that can automatically configure multi-task networks to fulfill diverse deployment constraints.

In this work, we propose a novel one-shot neural architecture search framework named AutoTaskFormer to tackle these challenges. We construct a sizeable search space based on Swin Transformer \cite{liu2021swin} that covers the most often-tuned hyper-parameters (\eg, embedding dimension, number of heads) in vision transformers. While this \textit{cell search space} for individual components covers a broad range of networks that can be searched, we determined through empirical observation that it is insufficient for scenarios where the primary challenge is to identify shared patterns among multiple tasks. In order to obtain a high-quality multi-task supernet with properly shared tasks, we advocate sampling multi-task subnets from the same multi-task conditional distribution in both supernet training and search phases. We refer to this multi-task conditional distribution search as \textit{skeleton search} which can be optimized along with the supernet training. To search multi-task subnets under different deployment requirements, we propose a novel criterion to distinguish the good ones from the bad. 

Our contributions are: (1) We propose a novel skeleton space integrated with a cell space to simultaneously search task-sharing patterns and detailed network architectures for multi-task learning. (2) We propose a novel one-shot multi-task NAS pipeline that automatically identifies multi-task networks to various deployment requirements without extra fine-tuning. (3) AutoTaskFormer achieves state-of-the-art multi-task learning performance on both small-scale (2-task Cityscape \cite{cordts2016cityscapes} and 3-task NYUv2 \cite{silberman2012indoor}) and large-scale (16-task Taskonomy \cite{zamir2018taskonomy} (shown in Fig.\ref{fig:pareto})) datasets, and can be easily adapted to new domains and tasks yielding better generalization than handcrafted transformers. 

\section{Related Work}

\textbf{Vision Transformers.} Our work is related to vision transformers. ViT \cite{dosovitskiy2020image} proposes the first design for image classification task by stacking transformer blocks on linear projections of non-overlapping image patches. Swin Transformer \cite{liu2021swin} introduces non-overlapping window partitions and restricts self-attention within each window. MViTv2 \cite{li2022mvitv2} incorporates decomposed relative positional embeddings and residual pooling connections to obtain a feature hierarchy from multiple stages. These works focus on designing general backbone networks that need human expertise to design task decoders when adapted to downstream tasks. In contrast, AutoTaskFormer is a framework that automates multi-task vision transformer design.

\textbf{Neural Architecture Search (NAS).} 
Our work is also related to NAS \cite{yu2020bignas, chen2021autoformer, chen2021searching}. Most recent efforts in NAS focus on one-shot weight-sharing strategy to amortize search cost \cite{tan2019efficientnet, guo2020single, brock2018smash}. The key idea is to train an over-parameterized supernet and share the weights among subnets \cite{yu2020bignas, guo2020single}. However, most existing NAS studies focus on searching network architectures for a single task. As we will explain in the next section, directly extending the one-shot weight-sharing strategy from single-task NAS to multi-task NAS incurs interferences among tasks, leading to significant performance degradation. AutoTaskFormer tackles this problem using a novel NAS pipeline that involves two search stages instead of one, specifically designed for multi-task learning.

\textbf{Multi-task Learning (MTL).} MTL aims to learn parameter sharing across multiple tasks. For example, Cross-Stitch \cite{misra2016cross} manually employs an additional group of shared units to merge multiple task backbones. Sluice \cite{ruder2019latent} jointly learns a latent multi-task architecture and task-specific models. NDDR-CNN \cite{gao2019nddr} proposes a novel CNN structure to learn a discriminative feature embedding for each task. 
There are several works that design task-sharing architectures through NAS. In particular, MTL-NAS \cite{gao2020mtl} starts with two task-specific networks and seeks optimal edges between the inter-task branches. AdaShare \cite{sun2020adashare} learns the sharing weights through a task-specific policy that selectively chooses which layers to execute for a given task. AutoMTL \cite{zhang2021automtl} proposes a source-to-source compiler that transforms the backbone CNN into a supermodel. However, due to the bottom-up approach of combining task-specific networks, the sharing between tasks and networks needs to be meticulously designed and hence can only support a small number of tasks. In contrast, AutoTaskFormer adopts a top-down approach that supports many more tasks with a novel NAS pipeline.

\begin{figure*}[t]
\begin{center}
\setlength{\tabcolsep}{9pt}
\begin{tabular}{c|c|c}
\includegraphics[height=3.2cm]{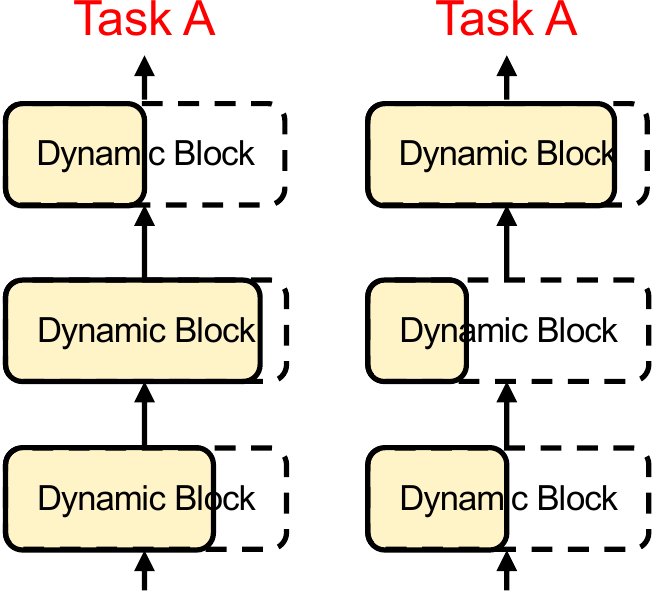} &
\includegraphics[height=3.2cm]{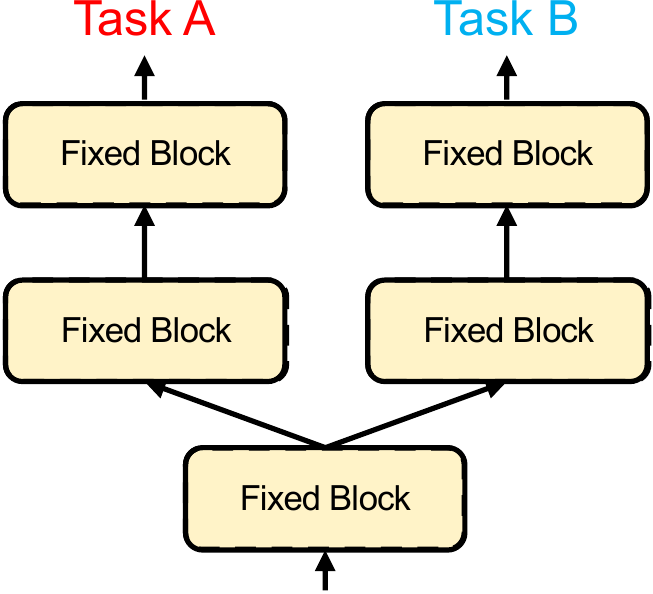} &
\includegraphics[height=3.2cm]{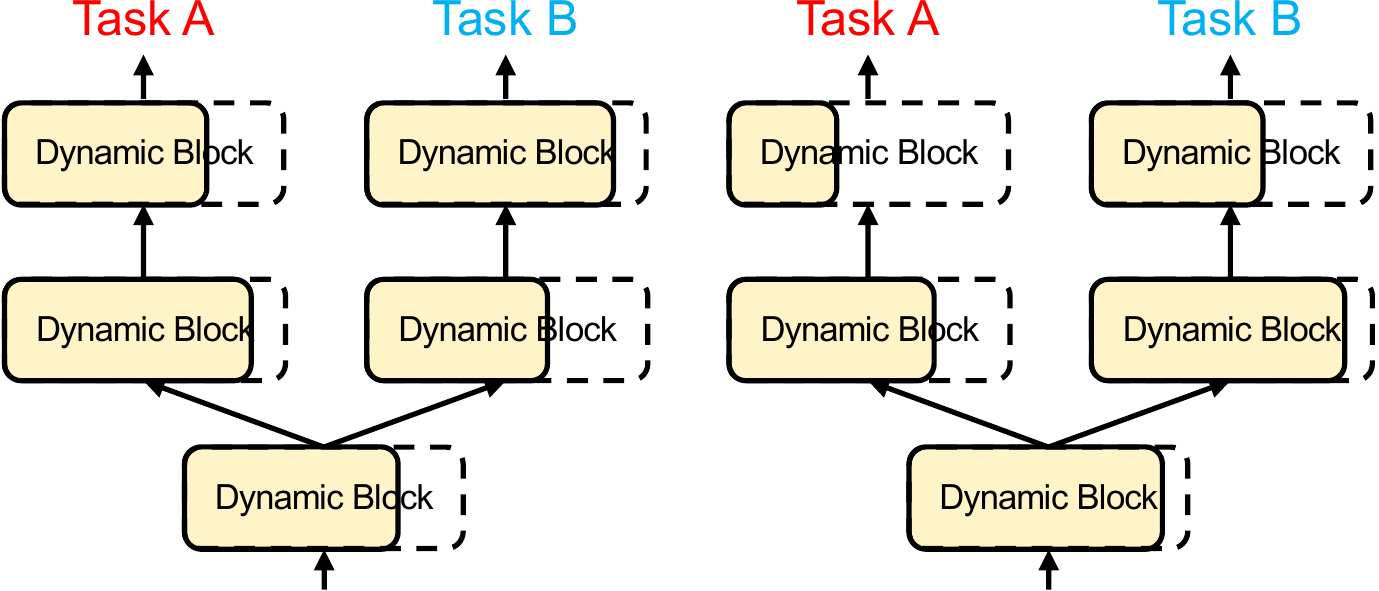} \\
\footnotesize{(a) Single-task NAS} & \footnotesize{(b) Multi-task Learning} & \footnotesize{(c) Our approach weight-shares subnets under distribution $\mathit{\Gamma}(\mathcal{A}~|~T_1, ..., T_n)$}
\end{tabular}
\end{center}
\caption{\textbf{(a)} Single-task NAS shares weights among various subnets in supernet training where subnets are sampled according to prior (uniform) distribution $\mathit{\Gamma}(\mathcal{A})$. \textbf{(b)} Multi-task learning shares part of the network weights for common knowledge. \textbf{(c)} Our approach encourages weight-sharing among multi-task subnets that are sampled from task conditional distribution $\mathit{\Gamma}(\mathcal{A}~|~T_1, ..., T_n)$. \label{fig:diff}}
\end{figure*}

\section{AutoTaskFormer}

\subsection{From Single-task NAS to Multi-task NAS\label{sec:3.1}}

To search a descent architecture within a vast space, a weight-sharing strategy is commonly leveraged to avoid exhausted subnet training from scratch \cite{yu2020bignas, chen2021autoformer, chen2021searching}. For a supernet $\mathcal{A}$ with weights $\mathcal{W}$, each subnet $\alpha$ directly inherits its weights $\mathcal{W}_\mathcal{A}(\alpha)$ from $\mathcal{W}$. The one-shot NAS is thus formulated as a two-stage optimization problem, \ie, supernet training and architecture searching. 

During supernet training, one or several subnets are sampled from $\mathcal{A}$ to directly update supernet weights $\mathcal{W}$ \cite{chen2021autoformer, chen2021searching} (as shown in Fig. \ref{fig:diff}(a)). By sharing weights among various subnets with weights inherited from $\mathcal{W}$, two-stage NAS makes every subnet $\alpha \in \mathcal{A}$ sufficiently trained such that they can be directly used for deployment without extra fine-tuning. To summarize, single-task NAS shares weights among various subnets during supernet training that minimize:
\begin{equation}
\mathcal{W}^* = \underset{\mathcal{W}}{\mathrm{argmin}}\ \mathbb{E}_{\alpha\sim \mathit{\Gamma}(\mathcal{A})}[\mathcal{L}(\mathcal{W}_\mathcal{A}(\alpha))],
\end{equation}
where $\mathit{\Gamma}(\mathcal{A})$ is a prior distribution of $\alpha\in \mathcal{A}$ and is empirically instantiated by uniform sampling \cite{guo2020single}.

\begin{figure}[t]
\begin{center}
\setlength{\tabcolsep}{1pt}
\newcommand{\centered}[1]{\begin{tabular}{l} #1 \end{tabular}}
\begin{tabular}{cc}
\includegraphics[width=0.23\textwidth]{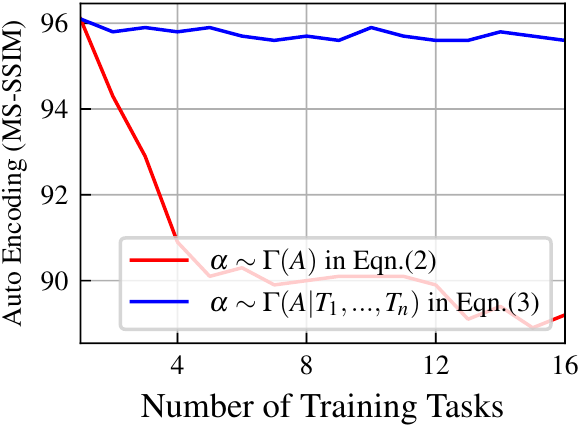} &
\includegraphics[width=0.23\textwidth]{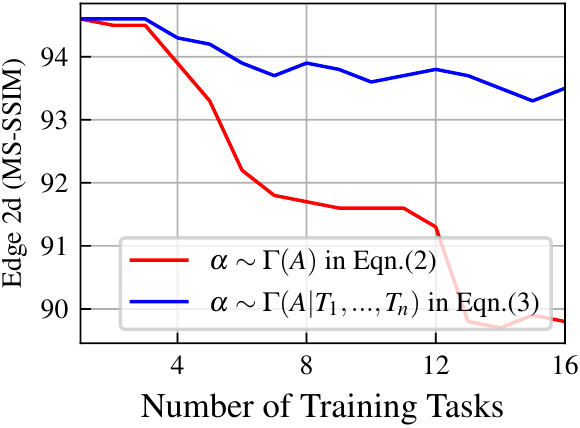} \\
\footnotesize{(a) \emph{Auto Encoding} performance} & \footnotesize{(b) \emph{Edge 2d} performance} \\
\multicolumn{2}{c}{\includegraphics[width=0.48\textwidth]{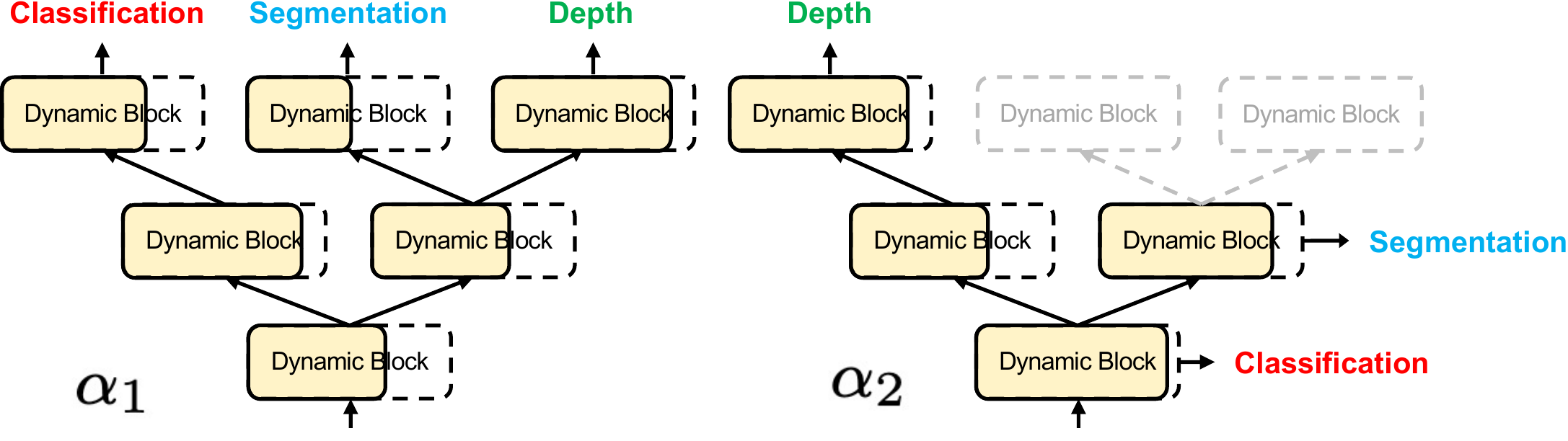}} \\
\multicolumn{2}{c}{\footnotesize{(c) Different task-sharing patterns when $\alpha_1$ and $\alpha_2$ sampled from $\mathit{\Gamma}(\mathcal{A})$}} \\
\end{tabular}
\end{center}
\vspace{-5pt}
\caption{\textbf{(a-b):} Two sharing strategies in supernet training for tasks \emph{Auto Encoding} and \emph{Edge 2d} (from Taskonomy\cite{zamir2018taskonomy}) respectively when the number of trained tasks increase. \textbf{(c):} Example two 3-task subnets $\alpha_1$, $\alpha_2$ sampled from the uniform distribution $\mathit{\Gamma}(\mathcal{A})$. \label{fig:ab_macro}}
\end{figure}

In comparison, \textbf{for multi-task NAS where} $\alpha$ \textbf{denotes multi-task subnet and} \textbf{and} $\mathcal{A}$ \textbf{is its supernet} (see example multi-task subnets in Fig. \ref{fig:diff}(c)), a straightforward idea is to extend the single-task objective to a sum of multiple losses: 
\begin{equation}\label{eq:err}
\mathcal{W}^* = \underset{\mathcal{W}}{\mathrm{argmin}}\ \mathbb{E}_{\alpha \sim \mathit{\Gamma}(\mathcal{A})}\left[\displaystyle{\sum_{i=1}^{n}\lambda_i \mathcal{L}_i(\mathcal{W}_\mathcal{A}(\alpha))}\right],
\end{equation}
where $\lambda_i$ and $\mathcal{L}_i$ are the pre-defined task weight and loss metric for the $i$-th task, respectively. An early experiment, shown in Fig. \ref{fig:ab_macro}(a-b), shows that this simple extension has huge interference among different tasks.

Digging for deeper reasons why Eqn. (\ref{eq:err}) does not work well for multi-task NAS, we found that weight-sharing is also widely used in previous multi-task (without NAS) literature. However, their weight-sharing paradigm is quite different from NAS. As shown in Fig. \ref{fig:diff}(b), Instead of sharing weights among multiple candidate subnets in NAS, they share part of network blocks (\eg, the early stage of backbones) across different tasks. If properly shared, it can effectively avoid negative knowledge transfer~\cite{sun2020adashare, guo2020learning}. It is also shown in the literature \cite{standley2020tasks} that if the sharing is done erroneously (\eg, sharing among irrelevant tasks), multi-task learning will produce results even worse than single-task learning.

\begin{figure*}[t]
    \begin{center}
     \includegraphics[width=0.9\textwidth]{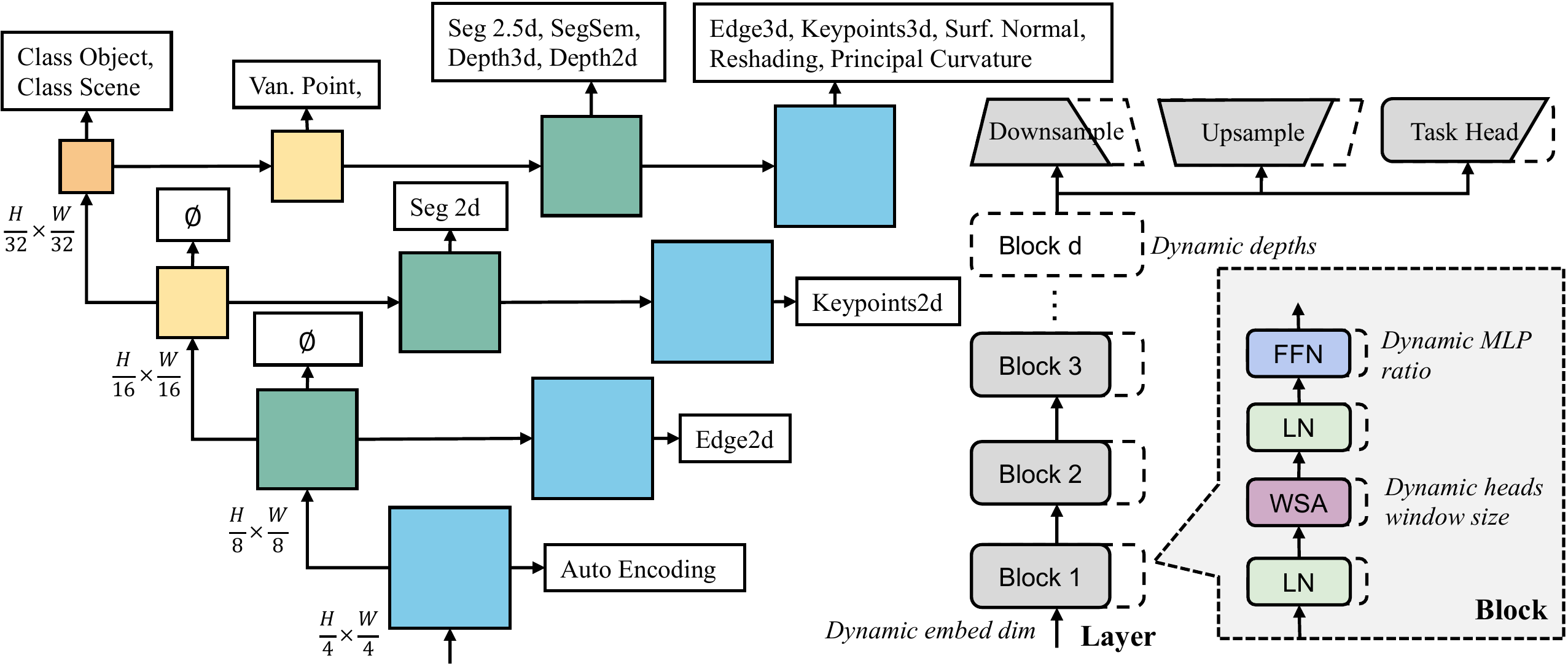}
     \end{center}
     \caption{\textbf{Left}: The skeleton space of AutoTaskFormer supernet where tasks can be attached at any adhoc point of middle layers. We show an example of searched vision transformer where $\varnothing$ indicates no task is attached at that layer. \textbf{Right}: The cell space where we search the optimal embedding dimension and depths for each layer, the number of heads, MLP ratio, and window size for each transformer block.}
     \label{fig:arch}
\end{figure*}

Since the weight-sharing among multiple tasks is not explicitly considered in Eqn. (\ref{eq:err}), when supernet training, both proper and erroneous multi-task subnets might be sampled. We illustrate with a simple example in Fig. \ref{fig:ab_macro} (c). Assuming $\alpha_2$ here is a low-quality subnet (\eg, predicting highly semantic classification with low-level features) that fails to converge, it will disturb the training of $\alpha_1$ due to the subnets' weight-sharing in NAS, thus ruining the whole training. To obtain a high-quality supernet for multi-task NAS, it is crucial to ensure that all subnets are qualified. In other words, \emph{we need a search space for multi-task supernet that we can sample well-qualified subnets under the same task-sharing pattern}. This is expressed as 
\begin{equation}
\resizebox{0.9\linewidth}{!}{%
$\mathcal{W}^* = \underset{\mathcal{W}}{\mathrm{argmin}}\ \mathbb{E}_{\alpha \sim \mathit{\Gamma}(\mathcal{A}~|~T_1,\ldots,T_n)}\left[\displaystyle{\sum_{i=1}^{n}\lambda_i \mathcal{L}_i(\mathcal{W}_\mathcal{A}(\alpha))}\right],$
}
\end{equation}
where $\mathit{\Gamma}(\mathcal{A}~|~T_1,...,T_n)$ is a conditional distribution of $\alpha \in \mathcal{A}$ with respect to task $T_1, ..., T_n$.

In the rest of this section, we first review the background of vision transformers. We then introduce how to get $\mathit{\Gamma}(\mathcal{A}~|~T_1,...,T_n)$, which we refer to as skeleton search. Finally, we elaborate our whole pipeline for multi-task NAS.

\subsection{Encoder-Decoder based Vision Transformer\label{sec:review}}

We review the encoder-decoder structure with window attention \cite{liu2021swin}, which serves as a basis of AutoTaskFormer.

During encoding, a 2D image is first transformed into patch embeddings by linear projection and then fed into the transformer layers. Fig. \ref{fig:arch} (Right) shows that a transformer layer consists of alternating blocks of window-partitioned multi-head self-attention (WSA) and feed-forward network (FFN). Layer normalization (LN) is applied before and after the residual connection of every block. 

During decoding, an upsampling layer is attached before each transformer layer to double the spatial resolution and halve the channel dimension. The decoder exploits semantic information at different stride levels and gradually recovers information in a coarse-to-fine manner.

\subsection{Search Space\label{sec:space}}
\paragraph{(Macro) Skeleton search space.} Our AutoTaskFormer is constructed on the basis of the encoder-decoder Swin Transformer, as illustrated in Fig. \ref{fig:arch} (Left). We extend the encoder by appending additional decoding layers at the middle levels. This architectural alteration introduces minimal parameters, as the majority of the parameters are concentrated at layers of size 1/16 and 1/32 \cite{liu2021swin}. The branched transformer design increases the search space by generating multiple multiscale features from different levels of the encoder.

\begin{figure}[t]
\begin{center}
\setlength{\tabcolsep}{4pt}
\begin{tabular}{cc}
\includegraphics[width=0.22\textwidth]{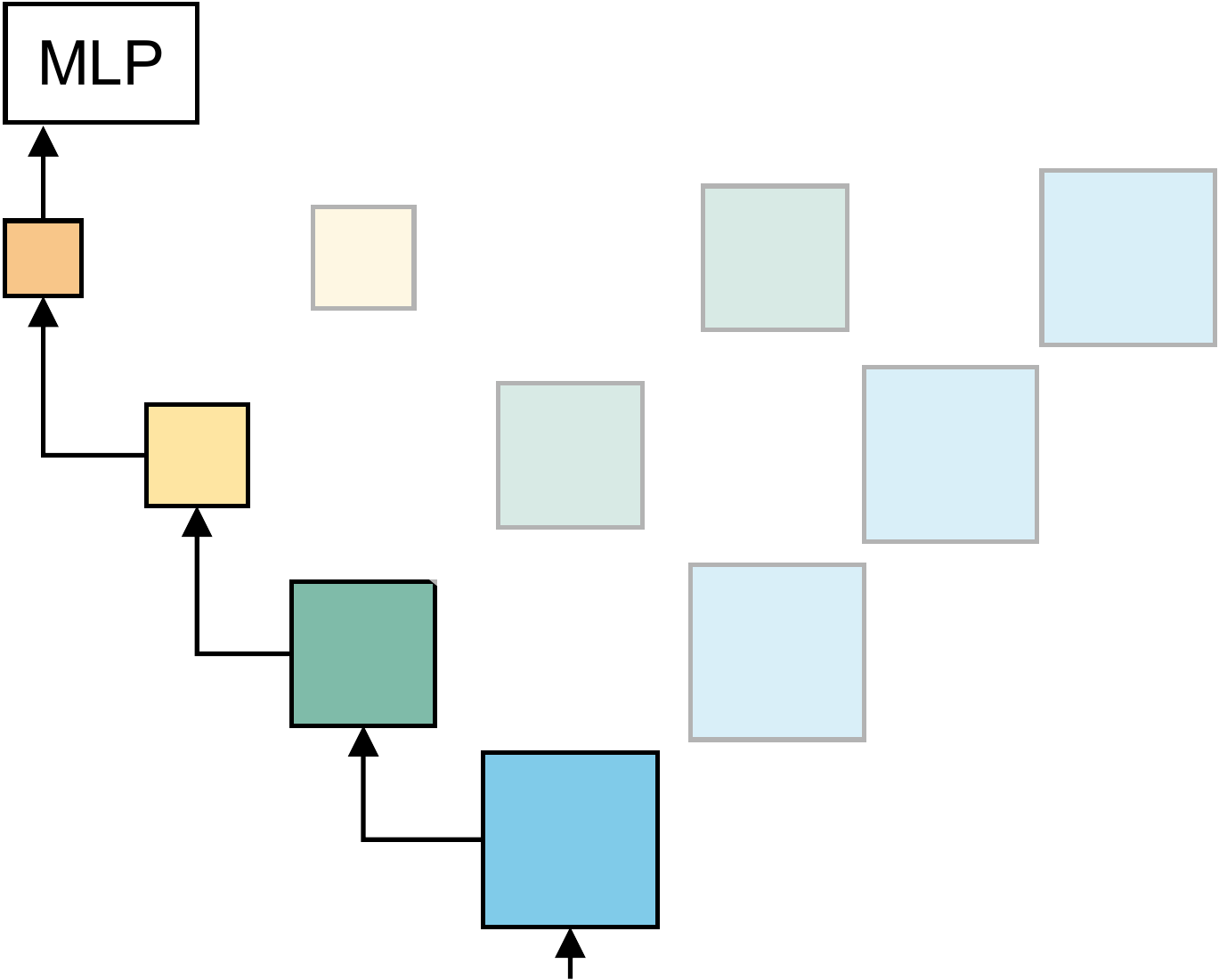} &
\includegraphics[width=0.22\textwidth]{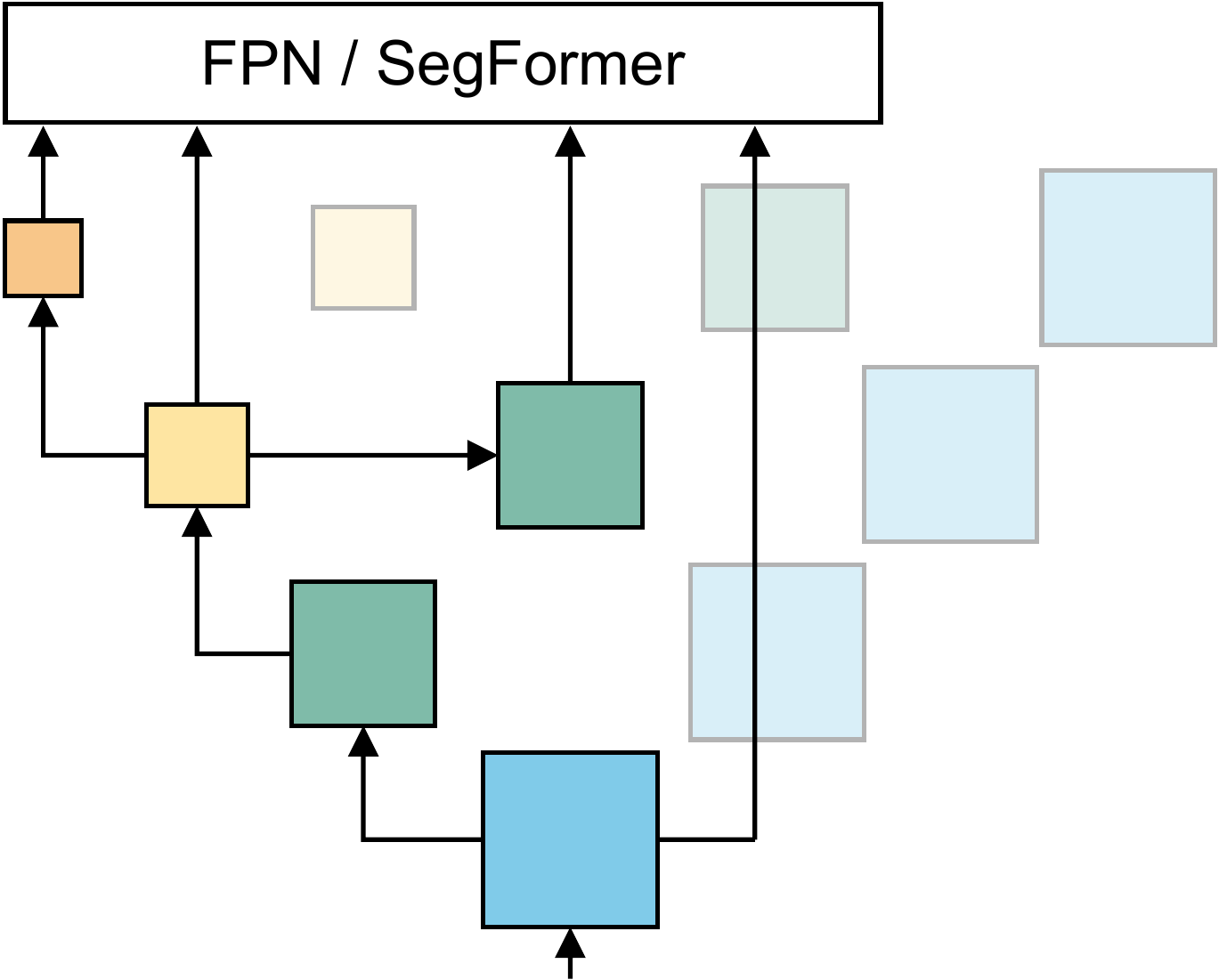} \\
\footnotesize{Example single-scale skeleton} & \footnotesize{Example multi-scale skeleton}
\vspace{-5pt}
\end{tabular}
\end{center}
\caption{\textbf{Left:} Single-scale skeletons cover existing designs like Swin \cite{liu2021swin}. \textbf{Right:} Multi-scale skeletons produce a multi-scale feature hierarchy for FPN \cite{lin2017feature} or SegFormer \cite{xie2021segformer}.\label{fig:skeleton}}
\end{figure}

We define a \textbf{skeleton} to be a single acyclic path. Fig. \ref{fig:skeleton} illustrates that the skeleton can either be single-scale, generating a single feature map for downstream tasks, or multi-scale, producing a feature hierarchy of sizes 1/4, 1/8, 1/16, and 1/32. For example, the single-scale skeleton space, depicted in Fig. \ref{fig:skeleton} (Left), includes the decoder-free Swin \cite{liu2021swin}. On the other hand, the multi-scale skeleton space, shown in Fig. \ref{fig:skeleton} (Right), is versatile enough to cater to multi-scale task heads, such as FPN \cite{lin2017feature} and SegFormer \cite{xie2021segformer}.

The skeleton space (either single-scale or multi-scale) is pre-defined by users according to their requests. In Fig. \ref{fig:arch} (Left), we show that AutoTaskFormer has $\{$\sqbox{corange} $\times 1$, \sqbox{cyellow}$\times 2$, \sqbox{cgreen}$\times 3$, \sqbox{cblue}$\times 4$ $\}$ transformer layers to obtain a feature hierarchy of size 1/32, 1/16, 1/8, and 1/4, respectively. We allow output features from any transformer layers for single-scale skeleton, resulting in $10=C_{4+3+2+1}^1$ searchable skeletons for each task. For multi-scale, we allow features from any combination of transformer layers with different resolutions, resulting in $24=C_4^1C_3^1C_2^1C_1^1$ candidates for each task.

Though the number of candidate skeletons for each task is limited, the skeleton space will exponentially increase according to the number of training tasks. For example, in 16 tasks learning, it contains $10^{16}$ and $24^{16}$ candidate skeletons in single-scale and multi-scale space, respectively.

\paragraph{(Micro) Cell search space.} We define a \textbf{cell} to be a single transformer layer consisting of a dynamic number of transformer blocks. As shown in Fig. \ref{fig:arch} (Right), each block follows the design of the Swin Transformer but with a searchable number of heads, MLP ratio, and window size. Since the input and output of a transformer block are connected via a shortcut connection, the embedding dimensions of input and output tensors should be consistent across all transformer blocks within the same cell. Thus, embedding dimension and depths are the cell-wise searchable parameters, while window size, number of heads, and MLP ratio are the block-wise searchable parameters. 

Following the weight sharing paradigm \cite{chen2021autoformer}, we encode the cell search space into a supernet. Every model in the search space is a subnet of the supernet, and all subnets share weights of their common parts. The supernet is the largest model in that space that owns the largest number of transformer blocks, the number of heads, embedding dimension, and MLP ratio within each transformer layer. According to the constraints on model parameters, we partition the cell space following the designs of Swin-S and Swin-B. The detailed cell search space is described in Appendix \ref{sec:cell}.

\subsection{Search Algorithm and Multi-task NAS Pipeline\label{sec:alg}}
\paragraph{Skeleton search algorithm.} We propose to simultaneously search skeletons for each task in multi-task supernet training such that tasks will share certain stages of knowledge according to their favored skeletons. The skeleton search algorithm is detailed as follows:

For the $s$-th skeleton in skeleton space $\mathbf{S}$, we seek a binary variable $u_{s,k}$ to determine whether that skeleton is searched to execute the $k$-th task $T_k \in \mathbf{T}$, yielding the best overall performance for all tasks. The selection among all candidate skeletons for all tasks is denoted by $ \mathbf{U}=\{u_{s,k}\}$, where each column is a one-hot vector.

To jointly optimize $\mathbf{U}$ along with network parameters through back-propagation, we adopt the Gumbel-softmax trick \cite{jang2017categorical}. Concretely, We let $\pi_k \in \mathbb{R}^{|\mathbf{S}|}$ be the distribution vector of the $k$-th column of $\mathbf{U}$ that we want to optimize, where the $\pi_{k,s}$ represents the probability that the $s$-th skeleton is selected to execute in the $k$-th task. We relax column-wise one-hot $\mathbf{U}$ to soft $\mathbf{U}'=\{u'_{s,k}\}$ by the reparameterization trick \cite{jang2017categorical}:
\begin{equation}\label{eqn:u}
    u'_{s,k}=\frac{\mathrm{exp}((\mathrm{log}\pi_{k,s} + \epsilon_s)/\tau)}{\sum_{s'\in \mathbf{S}} \mathrm{exp}((\mathrm{log}\pi_{k,s'} + \epsilon_{s'})/\tau)},
\end{equation}
where $\epsilon \in \mathbb{R}^{|\mathbf{S}|}$ is a vector with i.i.d. samples drawn from Gumbel distribution (0, 1), and $\tau$ is the softmax temperature. We initialize $\tau=5$ and gradually anneal it down to $0$ during training. For inference, we discretize $\mathbf{U}'$ to get $\mathbf{U}$ by $\mathrm{argmax}$, which is the preferred skeleton of each task.

\paragraph{Multi-task NAS pipeline.} Our multi-task NAS also includes train and search stages like conventional single-task NAS \cite{yu2020bignas, chen2021autoformer, chen2021searching}. We detail the multi-task NAS pipeline as follows (see pseudo-codes in Appendix \ref{sec:pseudo}): 

\textit{Train Stage: Multi-task Supernet Training.} In each iteration, we uniformly sample the largest, the smallest, and two middle-sized subnets from the cell space. For each sampled subnet, we calculate all task losses from all candidate skeletons in a loss matrix $\mathcal{L}\in \mathbb{R}^{|\mathbf{S}| \times |\mathbf{T}|}$, where $|\cdot|$ indicates the cardinality of the set. We use $\mathbf{U}'$ in Eqn. (\ref{eqn:u}) to weighted accumulate losses with respect to all candidate skeletons for each task, and the final loss $l$ is defined as:
\begin{equation}\label{eq:l}
    l = \sum_k\sum_{s}\lambda_k u'_{s, k} * \mathcal{L}_{s, k},
\end{equation}
where $\lambda_k$ is the pre-defined weight for the $k$-th task included in Appendix \ref{sec:detail_datasets}. We use $l$ to update both supernet parameters and skeleton distributions $\mathbf{U}'$.

\textit{Search Stage 1: Skeleton Space Search.} 
We perform the skeleton search before the cell space search to obtain optimal network structures for $\mathbf{T}$ tasks. Specifically, we first obtain discrete $\mathbf{U}$ from $\mathbf{U}'$ by $\mathrm{argmax}$ to decide the favored skeleton for each task. Then, we union the skeletons for all tasks in $\mathbf{T}$ to find the least common parts of layers, such that we can perform those tasks concurrently. Let's take two skeletons (2-task) as an example: \texttt{b1->head1} for task $T_1$ and \texttt{b1->pool1->b2->head2} for task $T_2$. We break down each skeleton into a set of components (\eg, $T_1$ consists of $\{$\texttt{b1}, \texttt{head1}$\}$). Afterward, we take the union of these two sets to obtain the multi-task (2-task) architecture.

\textit{Search Stage 2: Cell Space Search.}
To enable searches among thousands of subnets in $\mathcal{A}$ to perform $\mathbf{T}$ tasks concurrently, we first define a criterion to distinguish good multi-task subnets from bad ones. Motivated by \cite{maninis2019attentive, sun2020adashare}, for subnet $\alpha \in \mathcal{A}$ on task $T_k \in \mathbf{T}$, we use $\gamma(T_k; \alpha)$ to represent the relative multi-task performance for subnet $\alpha$ with respect to the averaged performances by:
\begin{equation}\label{eq:rel}
    \gamma(T_k; \alpha) = \frac{1}{|M|} \sum_{j=0}^{|M|} (-1)^{m_{j}} \frac{M_{\alpha,j} - \frac{1}{|\mathcal{A}|}\sum_{\alpha'} M_{\alpha',j}}{\frac{1}{|\mathcal{A}|}\sum_{\alpha'} M_{\alpha',j}},
\end{equation}
where $|M|$ is the number of metrics of the task $T_k$ and $m_{j}$ is 1 if a lower value represents a better performance for $j$-th metric and 0 otherwise. $M_{\alpha,j}$ is the value of $j$-th metric in task $T_k$ for subnet $\alpha$. We then define $\gamma (\alpha)$ to be the average relative performance for subnet $\alpha$:
\begin{equation}\label{eq:gamma}
\gamma(\alpha)= \frac{1}{|\mathbf{T}|}\sum_{T_k\in \mathbf{T}} \gamma (T_k; \alpha).
\end{equation}

We search for subnets $\alpha \in \mathcal{A}$ according to $\gamma(\alpha)$ that yield the best multi-task performance under different resource constraints via the evolution search \cite{chen2021autoformer, guo2020single} in Appendix \ref{sec:evolution_search}.

\begin{table*}[t]
\begin{center}
\small
\setlength{\tabcolsep}{3pt}
\aboverulesep=0.5ex
\belowrulesep=0.5ex
\resizebox{\textwidth}{!}{%
\renewcommand\arraystretch{1.1}
\newcommand{\CC}{\cellcolor{Gray}}
\begin{tabular}{l|c|ccc|ccccccccccccc|c}
\toprule
\multicolumn{1}{c|}{} & \multicolumn{1}{c|}{} & \multicolumn{3}{c|}{\textbf{Point Predictions}}  & \multicolumn{13}{c|}{\textbf{Dense Predictions}} & \\
\textbf{Models} & \textbf{\#Params} & \multicolumn{2}{c}{\footnotesize{Acc.$\uparrow$}} & \footnotesize{Loss$\downarrow$} & \footnotesize{mIoU$\uparrow$} & \multicolumn{10}{c}{\footnotesize{MS-SSIM $\uparrow$}} & \multicolumn{2}{c|}{\footnotesize{Loss $\downarrow$}}  & \multirow{2}{*}{$\Delta_T\uparrow$} \\
\cmidrule(lr){3-4} \cmidrule(lr){5-5} \cmidrule(lr){6-6} \cmidrule(lr){7-16} \cmidrule(lr){17-18} 
\multicolumn{1}{c|}{} & \multicolumn{1}{c|}{} & \footnotesize{Obj.} & \footnotesize{Scene} & \footnotesize{Van. Pt} & \footnotesize{SegSem} & \footnotesize{$\mathcal{D}$e} & \footnotesize{$\mathcal{D}$z} & \footnotesize{\textit{E}2d} & \footnotesize{\textit{E}3d} & \footnotesize{$\mathcal{K}$2d} & \footnotesize{$\mathcal{K}$3d} & \footnotesize{$\mathcal{N}$} & \footnotesize{$\mathcal{PC}$} & \footnotesize{$\mathcal{R}$} & \footnotesize{\textit{AE}} & \footnotesize{Seg2d} & \footnotesize{Seg2.5d} & \\
\midrule
Swin-T & 48.5 & 52.4 & 63.5 & 1.376 & 41.3 & 87.8 & 87.8 & 89.7 & 73.1 & 91.4 & 55.5 & 71.6 & 66.2 & 68.7 & 95.1 & 0.368 & 0.129 & 4.7 \\
CSwin-T & 31.3 & 52.0 & 63.8 & 1.388 & 39.6 & 88.0 & 88.0 & 88.7 & 72.6 & 91.5 & 55.0 & 71.1 & 66.0 & 69.0 & 94.6 & 0.407 & 0.133 & 3.3 \\
MViTv2-T & 44.3 & 53.3 & 66.3 & \textbf{1.346} & 45.5 & 88.2 & 88.2 & 87.7 & 73.2 & 90.6 & 56.1 & 71.8 & 66.3 & 70.0 & 93.9 & 0.381 & 0.128 & 5.9 \\
AutoFormer & 38.6 & 51.9 & 63.8 & 1.608 & 36.2 & 87.9 & 87.9 & 92.6 & 73.8 & 93.9 & 55.8 & 72.6 & 66.4 & 70.6 & 94.1 & 0.362 & 0.130 & 3.4 \\
\CC \textbf{AutoTaskFormer} (Ours) & \CC 45.5 & \CC \textbf{53.8} & \CC \textbf{66.5} & \CC 1.367 & \CC \textbf{45.6} & \CC \textbf{88.6} & \CC \textbf{88.6} & \CC \textbf{93.9} & \CC \textbf{75.8} & \CC \textbf{95.4} & \CC \textbf{59.0} & \CC \textbf{75.3} & \CC \textbf{67.6} & \CC \textbf{73.2} & \CC \textbf{95.8} & \CC \textbf{0.358} & \CC \textbf{0.127} & \CC \textbf{8.4} \\
\midrule
Swin-S & 69.8 & 53.3 & 65.4 & 1.492 & 43.2 & 88.1 & 88.1 & 89.9 & 73.7 & 91.7 & 56.6 & 72.4 & 66.6 & 70.1 & 95.2 & 0.368 & 0.128 & 5.4 \\
CSwin-S & 43.7 & 51.9 & 64.5 & 1.630 & 40.6 & 88.3 & 88.3 & 93.5 & 74.7 & 94.3 & 57.2 & 73.9 & 66.9 & 72.3 & 88.6 & 0.430 & 0.129 & 3.7 \\
MViTv2-S & 55.0 & 53.5 & 66.3 & 1.461 & 45.4 & 88.3 & 88.3 & 88.6 & 73.7 & 91.3 & 56.9 & 72.3 & 66.5 & 70.7 & 94.3 & 0.395 & 0.130 & 5.6 \\
AutoFormer & 56.6 & 52.3 & 64.4 & 1.584 & 38.3 & 87.9 & 87.9 & 93.4 & 74.3 & 94.7 & 56.6 & 73.0 & 66.6 & 71.1 & 94.6 & 0.377 & 0.129 & 4.2 \\
\CC \textbf{AutoTaskFormer} (Ours) & \CC 52.2 & \CC \textbf{54.0} & \CC \textbf{66.8} & \CC \textbf{1.353} & \CC \textbf{45.9} & \CC \textbf{88.5} & \CC \textbf{88.5} & \CC \textbf{94.0} & \CC \textbf{76.0} & \CC \textbf{96.0} & \CC \textbf{59.1} & \CC \textbf{75.5} & \CC \textbf{67.7} & \CC \textbf{73.4} & \CC \textbf{95.8} & \CC \textbf{0.349} & \CC \textbf{0.126} & \CC \textbf{8.9} \\
\midrule
Swin-B & 107.7 & 53.4 & 65.0 & 1.397 & 42.7 & 88.1 & 88.2 & 90.6 & 74.0 & 92.5 & 56.7 & 72.7 & 66.6 & 70.3 & 95.4 & 0.375 & 0.129 & 5.7 \\
CSwin-B & 97.6 & 52.6 & 66.0 & 1.639 & 42.0 & 87.8 & 87.8 & 94.1 & 75.2 & 95.4 & 58.2 & 74.5 & 67.0 & 72.7 & 91.0 & 0.483 & 0.129 & 3.9 \\
MViTv2-B & 71.5 & 54.0 & \textbf{67.3} & 1.447 & 46.6 & 88.5 & 88.5 & 88.3 & 74.0 & 91.1 & 57.3 & 72.8 & 66.7 & 71.2 & 94.1 & 0.369 & 0.126 & 6.7 \\
AutoFormer & 67.6 & 52.5 & 64.7 & 1.616 & 39.2 & 87.9 & 87.9 & 93.5 & 74.5 & 94.8 & 57.1 & 73.3 & 66.8 & 71.4 & 94.6 & 0.383 & 0.130 & 4.4 \\
\CC \textbf{AutoTaskFormer} (Ours) & \CC 65.1 & \CC \textbf{54.2} & \CC 67.0 & \CC \textbf{1.350} & \CC \textbf{47.2} & \CC \textbf{88.7} & \CC \textbf{88.7} & \CC \textbf{93.8} & \CC \textbf{76.4} & \CC \textbf{95.9} & \CC \textbf{59.8} & \CC \textbf{75.9} & \CC \textbf{67.9} & \CC \textbf{73.9} & \CC \textbf{95.9} & \CC \textbf{0.353} & \CC \textbf{0.125} & \CC \textbf{9.4} \\
\bottomrule
\end{tabular}
}
\end{center}
\caption{\textbf{Comparisons of 16-task learning on Taskonomy} with both transformer-based and NAS-based models. Van. Pt's $l_2$ loss is multiplied by a factor of 100 for better readability. The highest performance is marked in \textbf{bold} in each category. \label{tab:taskonomy}}
\end{table*}

\section{Experiments}

\subsection{AutoTaskFormer for Taskonomy\label{sec:4.1}}

\paragraph{Tasks.} To conduct a comprehensive evaluation, we incorporated 11 additional tasks from the original Taskonomy dataset \cite{zamir2018taskonomy}, beyond the 5 dense prediction tasks assessed in previous works \cite{sun2020adashare, zhang2021automtl}. This extension allows us to cover a wider range of task types that include \textbf{3 point prediction tasks}: object classification (Obj.), scene classification (Scene), and vanishing point regression (Van. Pt), as well as \textbf{13 dense prediction tasks:} depth euclidean ($\mathcal{D}$e), depth zbuffer ($\mathcal{D}$z), edge texture (\textit{E}2d), edge occlusion (\textit{E}3d), 2D keypoint detection ($\mathcal{K}$2d), 3D keypoint detection ($\mathcal{K}$3d), surface normal ($\mathcal{N}$), principal curvatures ($\mathcal{PC}$), reshading ($\mathcal{R}$), auto encoding (\textit{AE}), unsupervised 2d segmentation (Seg2d), unsupervised 2.5d segmentation (Seg2.5d), and semantic segmentation (SegSem).

\paragraph{Metrics and Baselines.}
We use top-1 accuracy for classifications, mIoU for semantic segmentation, and MS-SSIM \cite{wang2004image} for other dense tasks. For unsupervised Seg2d, Seg2.5d and Van. Pt, we report their absolute losses. To compare the overall multi-task performance for different models, we use relative performance $\Delta_T$ \cite{maninis2019attentive} with respect to the single task baseline (Swin-T for Taskonomy). The calculation of $\Delta_T$ is equivalent to Eqn. (\ref{eq:rel}) except for replacing averaged subnets performance with single-task baseline performance. We use single-scale transformers followed by standard FCN \cite{long2015fully} heads for all dense predictions. For point predictions, we use global average pooling followed by a simple MLP.

\begin{figure}[t]
\begin{center}
\setlength{\tabcolsep}{5pt}
\begin{tabular}{cc}
\hspace{-10pt}\includegraphics[width=0.23\textwidth]{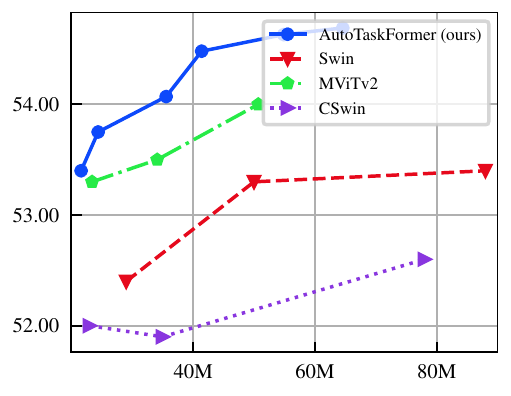} &
\includegraphics[width=0.23\textwidth]{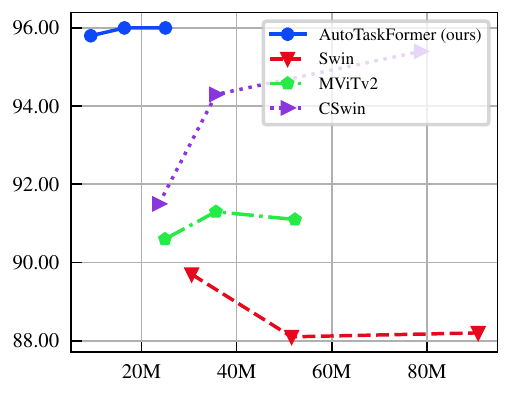} \\
\footnotesize{(a) Class Object (Acc., \%)} & \footnotesize{(b) Keypoints2d (MS-SSIM)}
\end{tabular}
\vspace{-5pt}
\end{center}
\caption{Performance of single-task models after multi-task training on Taskonomy. The other 14 tasks' models are in Appendix Fig.\ref{fig:supp_task_pareto}. \label{fig:task_pareto}}
\end{figure}

\begin{table*}[t]
\begin{center}
\small
\setlength{\tabcolsep}{7pt}
\renewcommand\arraystretch{1}
\newcommand{\CC}{\cellcolor{Gray}}
\begin{tabular}{clcccccccc}
\toprule
\multirow{3}{*}{\textbf{Model Type}} & \multirow{3}{*}{\textbf{Models}} & \multirow{3}{*}{\makecell{\textbf{\#Params} \\ \footnotesize{(M)}}} & \multicolumn{2}{c}{\textbf{Semantic Seg.}} & \multicolumn{5}{c}{\textbf{Depth Prediction}} \\
 &  &  &  \multirow{2}{*}{mIoU$\uparrow$} & \multirow{2}{*}{PAcc$\uparrow$} & \multicolumn{2}{c}{Error$\downarrow$} & \multicolumn{3}{c}{$\delta$ within$\uparrow$} \\
  \cmidrule(lr){6-7} \cmidrule(lr){8-10}
 &  &  &  &  & Abs & Rel & 1.25 & 1.25$^2$ & 1.25$^3$ \\
  \midrule
\multirow{3}{*}{\makecell{\textbf{CNN} \\ (\footnotesize{non-NAS MTL models})}}
 & Cross-Stitch \cite{misra2016cross} & 42.6 & 40.3 & 74.3 & 0.017 & 0.34 & 70.0 & 86.3 & 93.1 \\
  & Sluice \cite{ruder2019latent} & 42.6 & 39.8 & 74.2 & 0.018 & 0.35 & 68.9 & 85.8 & 92.8 \\
  & NDDR-CNN \cite{gao2019nddr} & 44.6 & 41.5 & 74.2 & 0.018 & 0.35 & 69.9 & 86.3 & 93.0 \\
 \midrule
\multirow{3}{*}{\makecell{\textbf{CNN} \\ (\footnotesize{NAS MTL models})}}
  & DEN \cite{ahn2019deep} & 23.8 & 38.0 & 74.2 & 0.017 & 0.37 & 72.3 & 87.1 & 93.4 \\
  & Adashare \cite{sun2020adashare} & 21.3 & 41.5 & 74.9 & 0.016 & 0.33 & 75.5 & 89.9 & 94.9 \\
  & AutoMTL \cite{zhang2021automtl} & 28.8 & 42.8 & 74.8 & 0.018 & 0.33 & 70.0 & 86.6 & 93.4 \\
 \midrule
\multirow{4}{*}{\makecell{\textbf{Transformer} \\ \footnotesize{(single-scale)}}} & Swin \cite{liu2021swin} & 30.6 & 49.4 & 91.3 & 0.015 & 0.28 & 76.2 & 90.3 & 95.4 \\
 & CSwin \cite{dong2022cswin} &  23.2 & 46.1 & 90.6 & 0.015 & 0.29 & 75.8 & 90.0 & 95.3 \\
 & MViTv2 \cite{li2022mvitv2} &  26.4 & 52.2 & 91.8 & 0.015 & \textbf{0.27} & 76.7 & 90.7 & 95.7 \\
 & \CC \textbf{AutoTaskFormer} (ours) & \CC 26.3 & \CC \textbf{54.7} & \CC \textbf{92.2} & \CC \textbf{0.013} & \CC \textbf{0.27} & \CC \textbf{77.7} & \CC \textbf{91.3} & \CC \textbf{96.0} \\
 \midrule
\multirow{4}{*}{\makecell{\textbf{Transformer} \\ \footnotesize{(multi-scale)}}} & Swin \cite{liu2021swin} & 28.8 & 51.3 & 91.5 & 0.017 & 0.30 & 74.6 & 89.2 & 95.0\\
 & CSwin \cite{dong2022cswin}  & 22.8 & 50.7 & 91.5 & 0.016 & 0.30 & 75.5 & 89.8 & 95.2 \\
 & MViTv2 \cite{li2022mvitv2}  & 24.6 & 55.1 & 92.5 & 0.015 & \textbf{0.27} & 76.4 & 90.6 & 95.9 \\
 & \CC \textbf{AutoTaskFormer} (ours)  & \CC 27.0 & \CC \textbf{56.8} & \CC \textbf{92.8} & \CC \textbf{0.012} & \CC \textbf{0.27} & \CC \textbf{78.8} & \CC \textbf{91.7} & \CC \textbf{96.1} \\
 \bottomrule
\end{tabular}
\end{center}
\caption{\textbf{Comparison of 2-task learning on Cityscapes} with both CNN-based and transformer-based methods. We use the tiny version of all baseline transformers to obtain similar parameters in CNN methods. The multi-scale transformer uses SegFormer task head \cite{xie2021segformer}\label{city}.}
\end{table*}

\paragraph{Results.} Tab. \ref{tab:taskonomy} shows that our AutoTaskFormers are consistently better than the baseline Swin Transformer \cite{liu2021swin} by 3.5$\sim$3.7\% on multi-task performances. AutoTaskFormer outperforms the state-of-the-art transformer models CSwin \cite{dong2022cswin} and MViTv2 \cite{li2022mvitv2} by 2.5\%$\sim$3.6\%. Regarding the absolute task performances, AutoTaskFormer surpasses baseline Swin and the latest MViTv2 in nearly all the tasks, illustrating the impact of multi-task NAS in designing and exploring architectures.

We compare the flexibility of NAS with other handcrafted vision transformers through the evaluation of our single-task subnet. As illustrated in Fig. \ref{fig:task_pareto}, the parameter range of our single-task subnet varies based on the tasks and their preferred skeletons. Thus we can directly sample a $\sim$5M parameter transformers (roughly three-quarters smaller than other handcrafted transformers) to perform 2D keypoint estimation without extra fine-tuning.

\subsection{AutoTaskFormer for Cityscapes\label{sec:city}}
\textbf{Cityscapes} \cite{cordts2016cityscapes} consists of high-resolution street-view images. We use this dataset for semantic segmentation and depth estimation tasks following \cite{liu2019end}. We evaluate both mIoU and pixel accuracy (PAcc) for segmentation. For depth estimation, we compute both absolute errors and the relative difference between the prediction and ground truth via the percentage of $\delta=\mathrm{max}\{\frac{y_\mathrm{pred}}{y_\mathrm{gt}}, \frac{y_\mathrm{gt}}{y_\mathrm{pred}}\}$ following \cite{eigen2014depth}.

We evaluate performances for both single-scale skeletons with MLP task heads and multi-scale skeletons with SegFormer \cite{xie2021segformer} heads. Tab. \ref{city} shows that AutoTaskFormer outperforms baseline Swin Transformer by a large margin of $\sim$4\% on semantic segmentation for both single-scale and multi-scale downstream task heads. Furthermore, it surpasses the most modern CSwin and MViTv2.

\begin{table}[t]
\begin{center}
\small
\setlength{\tabcolsep}{3pt}
\aboverulesep=0ex
\belowrulesep=0ex
\resizebox{0.5\textwidth}{!}{%
\renewcommand\arraystretch{1.05}
\begin{tabular}{l|c|cc|cccc|c}
\multirow{3}{*}{\textbf{Models}} & \multirow{3}{*}{\footnotesize{\makecell{\textbf{\#Params} \\ (M)}}}& \multicolumn{2}{c|}{\textbf{SemSeg.}} & \multicolumn{4}{c|}{\textbf{Surface Normal}} & \textbf{Depth} \\
 & & \multirow{2}{*}{\footnotesize{mIoU$\uparrow$}} & \multirow{2}{*}{\footnotesize{PAcc$\uparrow$}} & \multirow{2}{*}{\footnotesize{$\Delta\theta$ $\downarrow$}} & \multicolumn{3}{c|}{\footnotesize{$\Delta\theta$, within (\%) $\uparrow$}} & \multirow{2}{*}{\footnotesize{MS-SSIM$\uparrow$}} \\
 \cmidrule(lr){6-8}
 & & & & & \footnotesize{11.25$^{\circ}$} & \footnotesize{22.5$^{\circ}$} & \footnotesize{30$^{\circ}$} &  \\
 \midrule
Swin-S & 53.5 & 30.3 & 62.0 & 11.3 & 63.9 & 86.5 & 92.9 & 85.0 \\
CSwin-S & 36.2 & 27.1 & 59.5 & 11.0 & 65.6 & 86.8 & 92.8 & 86.7 \\
MViTv2-S & 38.6 & 32.7 & 63.7 & 10.9 & 65.5 & 87.0 & 93.1 & 87.1 \\
\textbf{AutoTaskFormer} & \textbf{35.3} & \textbf{33.2} & \textbf{64.4} & \textbf{10.0} & \textbf{69.4} & \textbf{88.3} & \textbf{93.7} & \textbf{88.0} \\
\end{tabular}
}
\end{center}
\caption{\textbf{Comparison of 3-task transfer learning on NYUv2} with other transformer-based models. All models are first pre-trained on Taskonomy and then fine-tuned on NYUv2.\label{tab:domain}}
\end{table}

\subsection{Transfer Learning\label{sec:transfer}}

\paragraph{Generalize to new domains.} \textbf{NYUv2} \cite{silberman2012indoor} comprises 1449 images from 464 different indoor scenes providing similar depth estimation, surface normal and semantic segmentation as in Taskonomy. We follow \cite{gao2019nddr} to evaluate multiple metrics for this dataset. Due to the scarcity of examples to train the vision transformer, we use this dataset to evaluate the transferability (\ie, attached new task heads and fine-tune the backbone models) to the unseen domain. Tab. \ref{tab:domain} shows that AutoTaskFormer yields better generalization when adapted to similar tasks under new domains. 

\begin{table}[t]
\begin{center}
\small
\setlength{\tabcolsep}{8pt}
\aboverulesep=0ex
\belowrulesep=0ex
\resizebox{0.4\textwidth}{!}{%
\renewcommand\arraystretch{1.05}
\begin{tabular}{l|ccc}
\multirow{2}{*}{\textbf{Models}} & \textbf{Jigsaw} & \textbf{Inpaint} & \textbf{Denoise} \\
 & \footnotesize{Acc.$\uparrow$} & \footnotesize{MS-SSIM $\uparrow$} & \footnotesize{MS-SSIM $\uparrow$} \\
 \midrule
Swin-S & 66.2 & 67.1 & 86.3 \\
CSwin-S & 50.7 & 69.4 & 87.9 \\
MViTv2-S & 64.4 & 73.1 & 92.3 \\
\textbf{AutoTaskFormer} & \textbf{72.0} & \textbf{73.2} & \textbf{93.0} \\
\end{tabular}
}
\end{center}
\caption{\textbf{Comparison of 3-task generalization on Taskonomy} with other transformer-based models. All models are pre-trained by 16 tasks on Taskonomy, with the frozen  backbone, and new task heads attached for generalization.\label{tab:task_adapt}}
\end{table}

\begin{figure*}[t]
\begin{center}
\setlength{\tabcolsep}{0pt}
\begin{tabular}{cccc}
\includegraphics[width=0.25\textwidth]{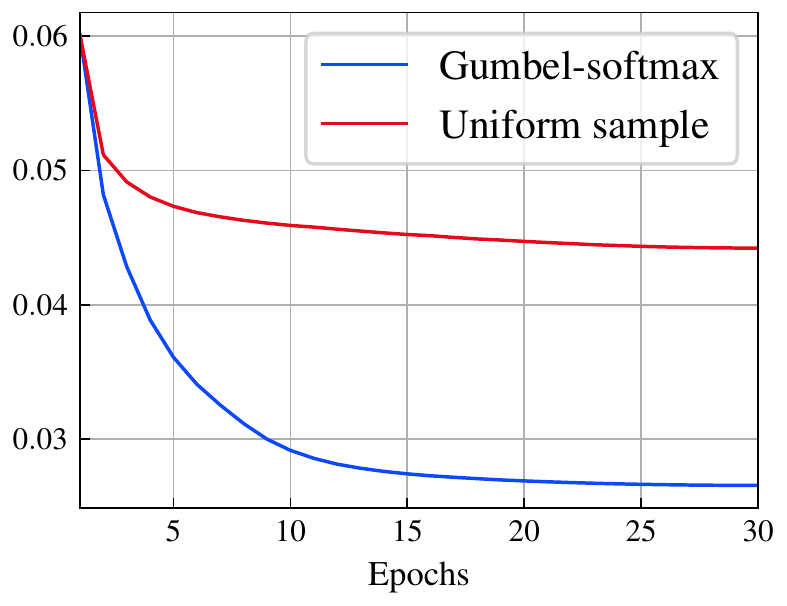} &
\includegraphics[width=0.25\textwidth]{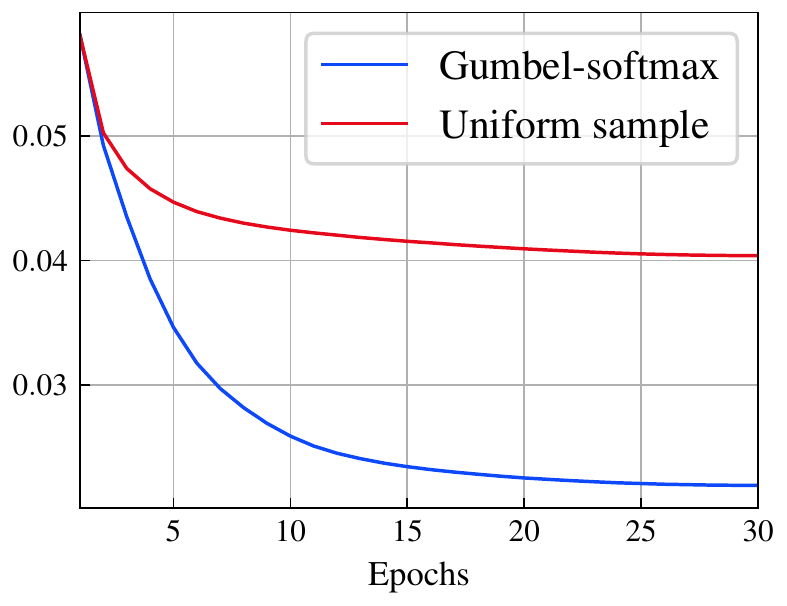} &
\includegraphics[width=0.25\textwidth]{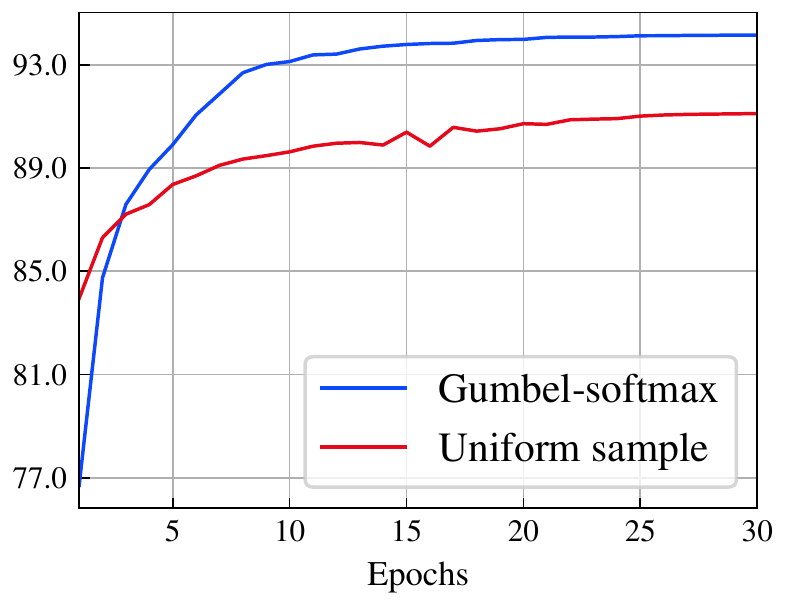} &
\includegraphics[width=0.25\textwidth]{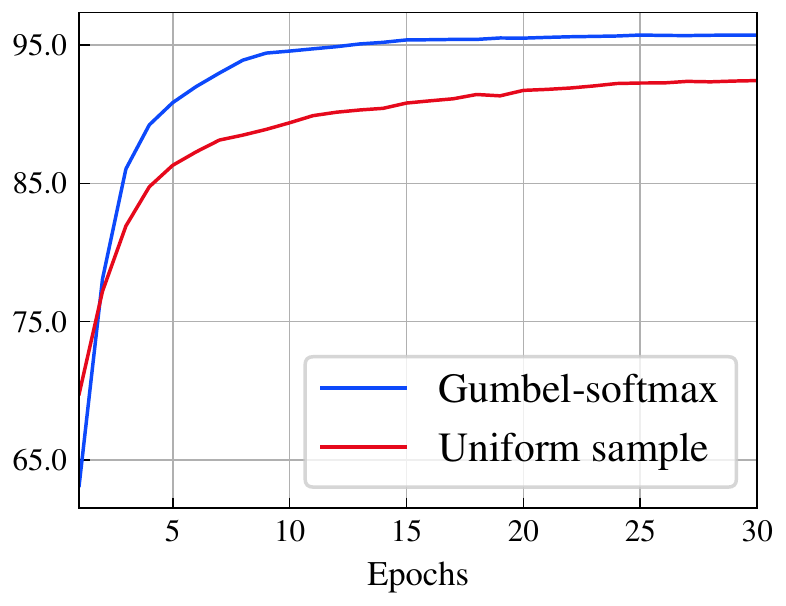} \\
\footnotesize{(a) Edge 2d (training loss)} & \footnotesize{(b) Keypoints2d (training loss)} & \footnotesize{(c) Edge 2d (test SSIM)} & \footnotesize{(d) Keypoints2d (test SSIM) }
\end{tabular}
\end{center}
\caption{Comparison between using and not using Gumbel-softmax to search task-favored skeletons in AutoTaskFormer: \textbf{(a-b)} compare the training loss between Gumbel-softmax and uniform sampling; \textbf{(c-d)} compare the validation performance during supernet training. \label{fig:ab_gumbel}}
\vspace{1mm}
\end{figure*}

\paragraph{Generalize to new tasks.} We also show how well AutoTaskFormer generalizes to new tasks on Taskonomy without fine-tuning the backbone. The transferred tasks include jigsaw puzzle, denoising, and image inpainting from \cite{zamir2018taskonomy}. In contrast to other training tasks in Sec. \ref{sec:4.1}, the novel tasks own completely different input styles (\ie, gaussian blurred, random puzzle permuted, or corrupted) and is thus challenging if we freeze the backbone network. The results in Tab. \ref{tab:task_adapt} show that AutoTaskFormer has better generalization performance when adapted to novel tasks under the same domain.

\subsection{Ablation Studies\label{sec:ablation}}

\textbf{Skeleton space \textit{vs}. SwinUNet.} As shown in Fig. \ref{fig:skeleton} (b), the largest skeleton in our single-scale framework is almost identical to SwinUNet (the baseline model in Tab. \ref{ab:comp}). As discussed in Sec. \ref{sec:3.1} and shown in Fig.  \ref{fig:ab_macro}, directly applying weight sharing as to joint train multi-task supernet is inappropriate. In contrast, AutoTaskFormer benefits from the tree structures that branched out at different levels of the encoder. We show quantitative results in Tab. \ref{ab:comp} that our single-scale skeleton space provides $\sim2\%$ relative improvements on Taskonomy. On the other head, our multi-scale skeleton space is about three times larger than SwinUNet that provides more optional skeletons for downstream multi-scale task heads like FPN \cite{zhao2017pyramid} and SegFormer \cite{xie2021segformer}.

\begin{table}[t]
\begin{center}
\small
\setlength{\tabcolsep}{3pt}
\resizebox{0.45\textwidth}{!}{%
\renewcommand\arraystretch{1.05}
\begin{tabular}{ccc|c|c}
\multicolumn{3}{c|}{\textbf{AutoTaskFormer}} & \multirow{2}{*}{\#params (M)} & \multirow{2}{*}{$\Delta_T$ (\%)} \\
\footnotesize{Skeleton space} & \footnotesize{Skeleton search} & \footnotesize{Cell space} & &  \\
\hlineB{2}
 \xmark & \xmark & \xmark & 41.0 & 4.4 \\
 \cmark & \xmark & \xmark & 43.5 & 6.3 \\
 \cmark & \cmark &  \xmark & 43.5 & 7.1 \\
 \cmark & \cmark & \cmark & 22.3 $-$ 108.7 & 7.0 $-$ 9.4 \\
\end{tabular}
}
\end{center}
\caption{Ablation studies of AutoTaskFormer on Taskonomy.\label{ab:comp}}
\end{table}

\textbf{Gumbel-softmax skeleton search.} AutoTaskFormer uses Gumbel-softmax \cite{jang2017categorical, sun2020adashare} to search the favored skeletons for each task during training and directly uses the searched skeleton for each task during the test. Another option is to uniformly sample skeletons during training and then search the task-favored skeleton according to its validation performance. In Fig. \ref{fig:ab_gumbel}, we compare these two methods on training loss and validation accuracy. It can be observed that the task training loss would be easily saturated at an early stage when we uniformly sample skeletons during training. A possible reason is that it would be impractical to expect less informative skeletons (\eg, a skeleton with a single transformer layer) to perform highly semantic predictions. 

In contrast, AutoTaskFormer uses Gumbel-softmax to search the favored skeleton for each task as presented in Sec. \ref{sec:alg}. Theoretically, there exists a negative correlation between $u'_{s,k}$ and $\mathcal{L}_{s,k}$ in Eqn. (\ref{eq:l}) when minimizing the overall loss $l$. It indicates that the $k$-th task gradually prefers the $s$-th skeleton with the minimum objective loss $\mathcal{L}_{s,k}$ and ignores the other candidate skeletons to prevent saturation. Fig. \ref{fig:ab_gumbel} (c-d) show that Gumbel-softmax search brought significant improvements for tasks like \textit{edge2d} and \textit{keypoints2d}.

\begin{figure}[t]
    \begin{center}
     \includegraphics[width=0.45\textwidth]{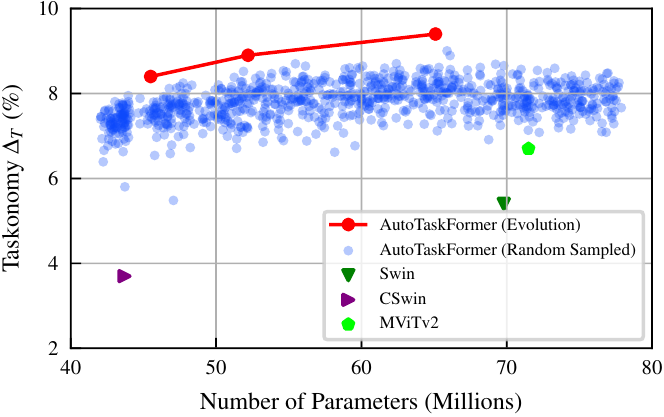}
     \end{center}
     \caption{Taskonomy performance of AutoTaskFormer and 1000 randomly sampled subnets within 40-80M parameters.\label{fig:cell}}
     \vspace{1mm}
\end{figure}

\textbf{Cell search space.} AutoTaskFormer adopts one-shot supernet training for multi-task NAS by weight sharing \cite{chen2021autoformer}. The blue points in Fig. \ref{fig:cell} show that there a large number of subnets achieving good performance when inheriting weights from the supernets. By adopting evolution search, AutoTaskFormer consistently achieves $\sim$0.5\% higher performances than the random search.

\section{Conclusion}

In this work, we present AutoTaskFormer, a novel one-shot neural architecture search framework for multi-task vision transformers. AutoTaskFormer not only  identifies the optimal skeleton for each task automatically but also adapts the cell designs in vision transformers to various resource constraints. AutoTaskFormer outperforms existing methods on multi-task benchmarks and exhibits promising generalization capabilities to unseen tasks and domains.

\clearpage

{\small
\bibliographystyle{ieee_fullname}
\bibliography{egbib}

\begin{thebibliography}{10}\itemsep=-1pt

\bibitem{ahn2019deep}
Chanho Ahn, Eunwoo Kim, and Songhwai Oh.
\newblock Deep elastic networks with model selection for multi-task learning.
\newblock In {\em Proceedings of the IEEE/CVF international conference on
  computer vision}, pages 6529--6538, 2019.

\bibitem{bhattacharjee2022mult}
Deblina Bhattacharjee, Tong Zhang, Sabine S{\"u}sstrunk, and Mathieu Salzmann.
\newblock Mult: An end-to-end multitask learning transformer.
\newblock In {\em Proceedings of the IEEE/CVF Conference on Computer Vision and
  Pattern Recognition}, pages 12031--12041, 2022.

\bibitem{brock2018smash}
Andrew Brock, Theo Lim, JM Ritchie, and Nick Weston.
\newblock Smash: One-shot model architecture search through hypernetworks.
\newblock In {\em International Conference on Learning Representations}, 2018.

\bibitem{cai2020once}
Han Cai, Chuang Gan, Tianzhe Wang, Zhekai Zhang, and Song Han.
\newblock Once for all: Train one network and specialize it for efficient
  deployment.
\newblock In {\em International Conference on Learning Representations}, 2020.

\bibitem{chen2021pre}
Hanting Chen, Yunhe Wang, Tianyu Guo, Chang Xu, Yiping Deng, Zhenhua Liu, Siwei
  Ma, Chunjing Xu, Chao Xu, and Wen Gao.
\newblock Pre-trained image processing transformer.
\newblock In {\em Proceedings of the IEEE/CVF Conference on Computer Vision and
  Pattern Recognition}, pages 12299--12310, 2021.

\bibitem{chen2021autoformer}
Minghao Chen, Houwen Peng, Jianlong Fu, and Haibin Ling.
\newblock Autoformer: Searching transformers for visual recognition.
\newblock In {\em Proceedings of the IEEE/CVF International Conference on
  Computer Vision}, pages 12270--12280, 2021.

\bibitem{chen2021searching}
Minghao Chen, Kan Wu, Bolin Ni, Houwen Peng, Bei Liu, Jianlong Fu, Hongyang
  Chao, and Haibin Ling.
\newblock Searching the search space of vision transformer.
\newblock {\em Advances in Neural Information Processing Systems},
  34:8714--8726, 2021.

\bibitem{cordts2016cityscapes}
Marius Cordts, Mohamed Omran, Sebastian Ramos, Timo Rehfeld, Markus Enzweiler,
  Rodrigo Benenson, Uwe Franke, Stefan Roth, and Bernt Schiele.
\newblock The cityscapes dataset for semantic urban scene understanding.
\newblock In {\em Proceedings of the IEEE conference on computer vision and
  pattern recognition}, pages 3213--3223, 2016.

\bibitem{dong2022cswin}
Xiaoyi Dong, Jianmin Bao, Dongdong Chen, Weiming Zhang, Nenghai Yu, Lu Yuan,
  Dong Chen, and Baining Guo.
\newblock Cswin transformer: A general vision transformer backbone with
  cross-shaped windows.
\newblock In {\em Proceedings of the IEEE/CVF Conference on Computer Vision and
  Pattern Recognition}, pages 12124--12134, 2022.

\bibitem{dosovitskiy2020image}
Alexey Dosovitskiy, Lucas Beyer, Alexander Kolesnikov, Dirk Weissenborn,
  Xiaohua Zhai, Thomas Unterthiner, Mostafa Dehghani, Matthias Minderer, Georg
  Heigold, Sylvain Gelly, et~al.
\newblock An image is worth 16x16 words: Transformers for image recognition at
  scale.
\newblock In {\em International Conference on Learning Representations}, 2020.

\bibitem{eigen2014depth}
David Eigen, Christian Puhrsch, and Rob Fergus.
\newblock Depth map prediction from a single image using a multi-scale deep
  network.
\newblock {\em Advances in neural information processing systems}, 27, 2014.

\bibitem{gao2020mtl}
Yuan Gao, Haoping Bai, Zequn Jie, Jiayi Ma, Kui Jia, and Wei Liu.
\newblock Mtl-nas: Task-agnostic neural architecture search towards
  general-purpose multi-task learning.
\newblock In {\em Proceedings of the IEEE/CVF Conference on computer vision and
  pattern recognition}, pages 11543--11552, 2020.

\bibitem{gao2019nddr}
Yuan Gao, Jiayi Ma, Mingbo Zhao, Wei Liu, and Alan~L Yuille.
\newblock Nddr-cnn: Layerwise feature fusing in multi-task cnns by neural
  discriminative dimensionality reduction.
\newblock In {\em Proceedings of the IEEE/CVF conference on computer vision and
  pattern recognition}, pages 3205--3214, 2019.

\bibitem{guo2020learning}
Pengsheng Guo, Chen-Yu Lee, and Daniel Ulbricht.
\newblock Learning to branch for multi-task learning.
\newblock In {\em International Conference on Machine Learning}, pages
  3854--3863. PMLR, 2020.

\bibitem{guo2020single}
Zichao Guo, Xiangyu Zhang, Haoyuan Mu, Wen Heng, Zechun Liu, Yichen Wei, and
  Jian Sun.
\newblock Single path one-shot neural architecture search with uniform
  sampling.
\newblock In {\em European conference on computer vision}, pages 544--560.
  Springer, 2020.

\bibitem{jang2017categorical}
Eric Jang, Shixiang Gu, and Ben Poole.
\newblock Categorical reparametrization with gumble-softmax.
\newblock In {\em International Conference on Learning Representations (ICLR
  2017)}. OpenReview. net, 2017.

\bibitem{li2022mvitv2}
Yanghao Li, Chao-Yuan Wu, Haoqi Fan, Karttikeya Mangalam, Bo Xiong, Jitendra
  Malik, and Christoph Feichtenhofer.
\newblock Mvitv2: Improved multiscale vision transformers for classification
  and detection.
\newblock In {\em Proceedings of the IEEE/CVF Conference on Computer Vision and
  Pattern Recognition}, pages 4804--4814, 2022.

\bibitem{lin2017feature}
Tsung-Yi Lin, Piotr Doll{\'a}r, Ross Girshick, Kaiming He, Bharath Hariharan,
  and Serge Belongie.
\newblock Feature pyramid networks for object detection.
\newblock In {\em Proceedings of the IEEE conference on computer vision and
  pattern recognition}, pages 2117--2125, 2017.

\bibitem{liu2019end}
Shikun Liu, Edward Johns, and Andrew~J Davison.
\newblock End-to-end multi-task learning with attention.
\newblock In {\em Proceedings of the IEEE/CVF conference on computer vision and
  pattern recognition}, pages 1871--1880, 2019.

\bibitem{liu2021swin}
Ze Liu, Yutong Lin, Yue Cao, Han Hu, Yixuan Wei, Zheng Zhang, Stephen Lin, and
  Baining Guo.
\newblock Swin transformer: Hierarchical vision transformer using shifted
  windows.
\newblock In {\em Proceedings of the IEEE/CVF International Conference on
  Computer Vision}, pages 10012--10022, 2021.

\bibitem{long2015fully}
Jonathan Long, Evan Shelhamer, and Trevor Darrell.
\newblock Fully convolutional networks for semantic segmentation.
\newblock In {\em Proceedings of the IEEE conference on computer vision and
  pattern recognition}, pages 3431--3440, 2015.

\bibitem{maninis2019attentive}
Kevis-Kokitsi Maninis, Ilija Radosavovic, and Iasonas Kokkinos.
\newblock Attentive single-tasking of multiple tasks.
\newblock In {\em Proceedings of the IEEE/CVF Conference on Computer Vision and
  Pattern Recognition}, pages 1851--1860, 2019.

\bibitem{misra2016cross}
Ishan Misra, Abhinav Shrivastava, Abhinav Gupta, and Martial Hebert.
\newblock Cross-stitch networks for multi-task learning.
\newblock In {\em Proceedings of the IEEE conference on computer vision and
  pattern recognition}, pages 3994--4003, 2016.

\bibitem{mohamed2021spatio}
Eslam Mohamed and Ahmad El~Sallab.
\newblock Spatio-temporal multi-task learning transformer for joint moving
  object detection and segmentation.
\newblock In {\em 2021 IEEE International Intelligent Transportation Systems
  Conference (ITSC)}, pages 1470--1475. IEEE, 2021.

\bibitem{ruder2019latent}
Sebastian Ruder, Joachim Bingel, Isabelle Augenstein, and Anders S{\o}gaard.
\newblock Latent multi-task architecture learning.
\newblock In {\em Proceedings of the AAAI Conference on Artificial
  Intelligence}, volume~33, pages 4822--4829, 2019.

\bibitem{seong2019video}
Hongje Seong, Junhyuk Hyun, and Euntai Kim.
\newblock Video multitask transformer network.
\newblock In {\em Proceedings of the IEEE/CVF International Conference on
  Computer Vision Workshops}, pages 0--0, 2019.

\bibitem{silberman2012indoor}
Nathan Silberman, Derek Hoiem, Pushmeet Kohli, and Rob Fergus.
\newblock Indoor segmentation and support inference from rgbd images.
\newblock In {\em European conference on computer vision}, pages 746--760.
  Springer, 2012.

\bibitem{standley2020tasks}
Trevor Standley, Amir Zamir, Dawn Chen, Leonidas Guibas, Jitendra Malik, and
  Silvio Savarese.
\newblock Which tasks should be learned together in multi-task learning?
\newblock In {\em International Conference on Machine Learning}, pages
  9120--9132. PMLR, 2020.

\bibitem{sun2020adashare}
Ximeng Sun, Rameswar Panda, Rogerio Feris, and Kate Saenko.
\newblock Adashare: Learning what to share for efficient deep multi-task
  learning.
\newblock {\em Advances in Neural Information Processing Systems},
  33:8728--8740, 2020.

\bibitem{tan2019efficientnet}
Mingxing Tan and Quoc~V Le.
\newblock Efficientnet: Rethinking model scaling for convolutional neural
  networks.
\newblock In {\em ICML}, 2019.

\bibitem{touvron2021training}
Hugo Touvron, Matthieu Cord, Matthijs Douze, Francisco Massa, Alexandre
  Sablayrolles, and Herv{\'e} J{\'e}gou.
\newblock Training data-efficient image transformers \& distillation through
  attention.
\newblock In {\em International Conference on Machine Learning}, pages
  10347--10357. PMLR, 2021.

\bibitem{wang2004image}
Zhou Wang, Alan~C Bovik, Hamid~R Sheikh, and Eero~P Simoncelli.
\newblock Image quality assessment: from error visibility to structural
  similarity.
\newblock {\em IEEE transactions on image processing}, 13(4):600--612, 2004.

\bibitem{xie2021segformer}
Enze Xie, Wenhai Wang, Zhiding Yu, Anima Anandkumar, Jose~M Alvarez, and Ping
  Luo.
\newblock Segformer: Simple and efficient design for semantic segmentation with
  transformers.
\newblock {\em Advances in Neural Information Processing Systems},
  34:12077--12090, 2021.

\bibitem{ye2022inverted}
Hanrong Ye and Dan Xu.
\newblock Inverted pyramid multi-task transformer for dense scene
  understanding.
\newblock In {\em Computer Vision--ECCV 2022: 17th European Conference, Tel
  Aviv, Israel, October 23--27, 2022, Proceedings, Part XXVII}, pages 514--530.
  Springer, 2022.

\bibitem{ye2023taskprompter}
Hanrong Ye and Dan Xu.
\newblock Taskprompter: Spatial-channel multi-task prompting for dense scene
  understanding.
\newblock In {\em The Eleventh International Conference on Learning
  Representations}, 2023.

\bibitem{yu2020bignas}
Jiahui Yu, Pengchong Jin, Hanxiao Liu, Gabriel Bender, Pieter-Jan Kindermans,
  Mingxing Tan, Thomas Huang, Xiaodan Song, Ruoming Pang, and Quoc Le.
\newblock Bignas: Scaling up neural architecture search with big single-stage
  models.
\newblock In {\em European Conference on Computer Vision}, pages 702--717.
  Springer, 2020.

\bibitem{zamir2018taskonomy}
Amir~R Zamir, Alexander Sax, William Shen, Leonidas~J Guibas, Jitendra Malik,
  and Silvio Savarese.
\newblock Taskonomy: Disentangling task transfer learning.
\newblock In {\em Proceedings of the IEEE conference on computer vision and
  pattern recognition}, pages 3712--3722, 2018.

\bibitem{zhang2021automtl}
Lijun Zhang, Xiao Liu, and Hui Guan.
\newblock Automtl: A programming framework for automating efficient multi-task
  learning, 2021.

\bibitem{zhao2017pyramid}
Hengshuang Zhao, Jianping Shi, Xiaojuan Qi, Xiaogang Wang, and Jiaya Jia.
\newblock Pyramid scene parsing network.
\newblock In {\em Proceedings of the IEEE conference on computer vision and
  pattern recognition}, pages 2881--2890, 2017.

\end{thebibliography}
}

\clearpage

\renewcommand\thesection{\Alph{section}}

\appendix

\section{Pseudo-code of Multi-task NAS Pipeline\label{sec:pseudo}}

\begin{algorithm}[!h]
    \caption{Multi-task NAS pipeline.}
    \label{alg:mel}
    \textbf{Notations:}
    \
    
    \hspace{1.5em} $\mathbf{T}$: all training tasks
    
    \hspace{1.5em} $\mathbf{D}$: all training data
    
    \hspace{1.5em} $\mathcal{A}$: supernet with skeleton space $\mathbf{S}$ and the weight $W$.
    
    \hspace{1.5em} $a$: subnet sampled from $\mathcal{A}$ with weights from $W$.
    
    
    \textbf{Multi-task Supernet Train}
    \
    \begin{algorithmic}[1]
    \FOR{\texttt{\_ in range(epochs)}}
        \FOR{\texttt{ $x, y(\mathbf{T})$ in {$\mathbf{D}$}}}
            \STATE $a \gets \textsc{RandomSample}(\mathcal{A})$
            \STATE Calculate $l$ according to Eqn.(\ref{eq:l})
            \STATE Update $\mathbf{U}'$ and $W$ by $l$.
        \ENDFOR
    \ENDFOR
    \end{algorithmic}

    \textbf{Skeleton Space Search}
    \
    \begin{algorithmic}[1]
        \STATE Get discrete $\mathbf{U}$ from $\mathbf{U}'$ by column-wise $\texttt{argmax}$.
        \STATE Get $|\mathbf{T}|$ task-favored skeletons $s_1, s_2, ..., s_{|\mathbf{T}|}$ by $\mathbf{U}$.
        \STATE Union skeletons to find the smallest skeleton $\hat{s}$ where every $s_i\subseteq \hat{s}$
    \end{algorithmic}

    \textbf{Cell Space Search}
    \
    \begin{algorithmic}[1]
        \STATE Search cell hyper-parameters for skeleton $\hat{s}$ according to Eqn.(\ref{eq:gamma}).
    \end{algorithmic}
\end{algorithm}

\section{Datasets and Implementation Details\label{sec:detail_datasets}}

\subsection{16 tasks learning on Taskonomy} Taskonomy \cite{zamir2018taskonomy} consists of 4.5 million images from 500 buildings with annotations available for 26 tasks. Considering the huge size of full dataset, we use the official tiny split from it, which consists of 260K images from 25 buildings for train, 60K images for validation, and 54K images for test. Unlike previous methods \cite{sun2020adashare, zhang2021automtl} that only evaluate 5 dense prediction tasks, we include 11 more tasks from the original 26 tasks  that covers more varieties of task types like classification, point regression, and unsupervised metric learning. Due to the huge capacity of this dataset, we directly resize all the input images to 224$\times$224 to train and test all vision transformers without any data augmentation. We follow the same loss criterion as in the original Taskonomy for all vision transformers.

For Taskonomy \cite{zamir2018taskonomy}, we train AutoTaskFormer as well as the other baseline vision transformers for 30 epochs. We adopt the sandwich rules \cite{yu2020bignas} that sample the largest, the smallest, and two middle sized subnets to train the supernet like AutoTaskFormer and AutoFormer \cite{chen2021autoformer} to aggregate their gradients in each iteration. We use AdamW optimization with a base learning rate of 5e-4 for batch size of 64. We use a linear warm-up strategy in the first epoch and a cosine scheduler to decrease learning rate to 5e-6. We set the weight decay to 0.05 for regularization. In Tab. \ref{tab:sup_taskonomy}, we provide task loss weights $\lambda_k$ and the baseline single-task learning benchmark to calculate the multi-task relative performance. We report the FLOPs of different multi-task vision transformers in Tab. \ref{tab:supp_flops}.

\begin{table}[t]
\begin{center}
\small
\setlength{\tabcolsep}{2pt}
\aboverulesep=0.5ex
\belowrulesep=0.5ex
\begin{tabular}{llccc}
\toprule
\multicolumn{2}{l}{\textbf{Models}} & \textbf{\#Params (M)} & \textbf{\#FLOPs (G)} & $\Delta_T$\\
\midrule
\multirow{3}{*}{Swin \cite{liu2021swin}} & Tiny & 48.5 & 13.4 & 4.7 \\
 & Small & 69.8 & 17.6 & 5.4 \\
 & Base & 107.7 & 24.2 & 5.7 \\
 \midrule
\multirow{3}{*}{CSwin \cite{dong2022cswin}} & Tiny & 31.3 & 8.1 & 3.3 \\
 & Small & 43.7 & 10.5 & 3.7 \\
 & Base & 97.6 & 23.4 & 3.9 \\
 \midrule
\multirow{3}{*}{MViTv2 \cite{li2022mvitv2}} & Tiny & 44.3 & 11.6 & 5.9 \\
 & Small & 55.0 & 16.1 & 5.6 \\
 & Base & 71.5 & 18.0 & 6.7 \\
 \midrule
\textbf{AutoTaskFormer} & \textbf{Auto} & \textbf{22.3$\sim$108.7} & \textbf{6.2$\sim$24.7} & \textbf{7.0$\sim$9.4}\\
\bottomrule
\end{tabular}
\end{center}
\caption{\textbf{FLOPs} of different vision transformers on Taskonomy.\label{tab:supp_flops}}
\end{table}

\begin{figure*}[t]
\begin{center}
\setlength{\tabcolsep}{10pt}
\begin{tabular}{cc}
\includegraphics[width=0.45\textwidth]{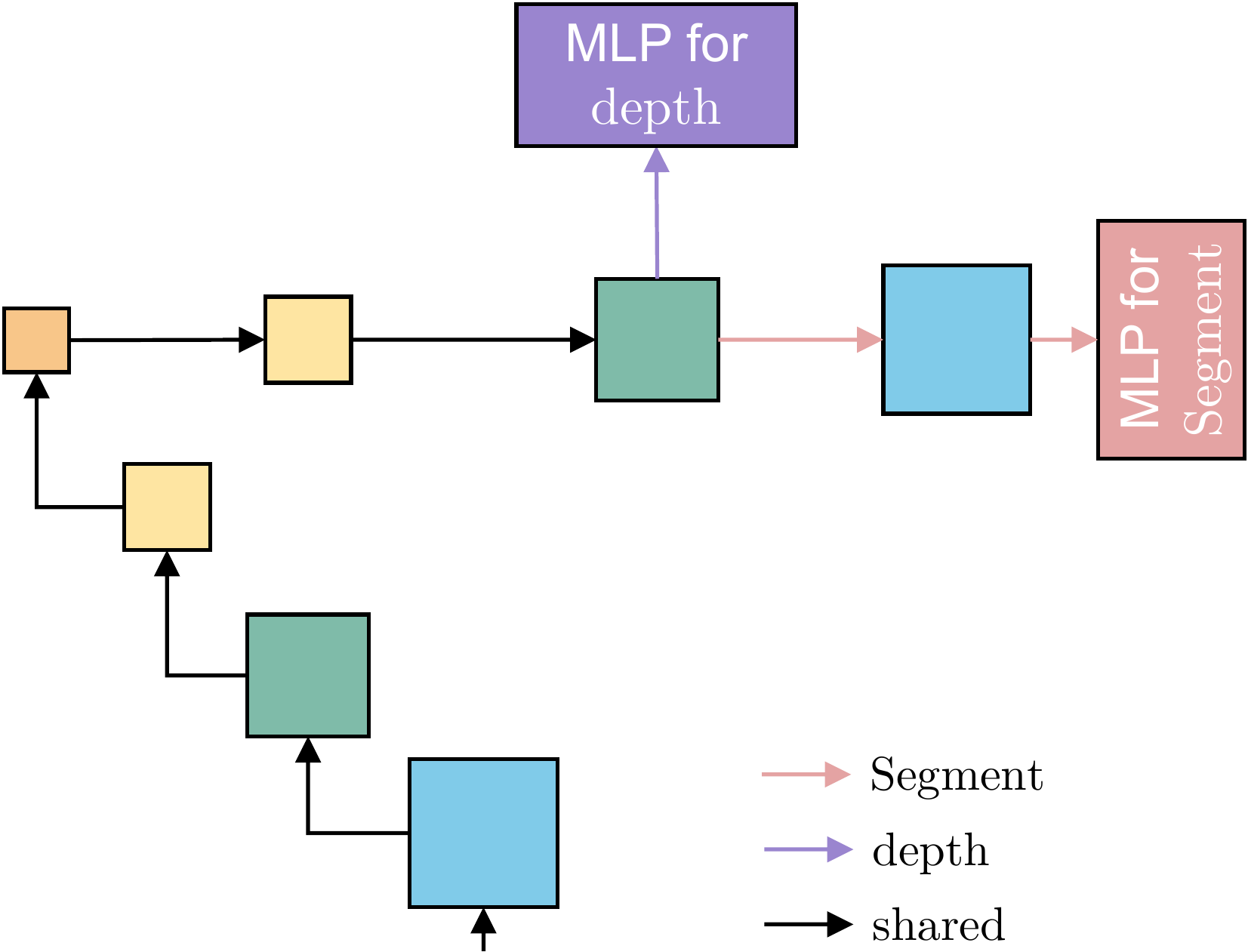} &
\includegraphics[width=0.45\textwidth]{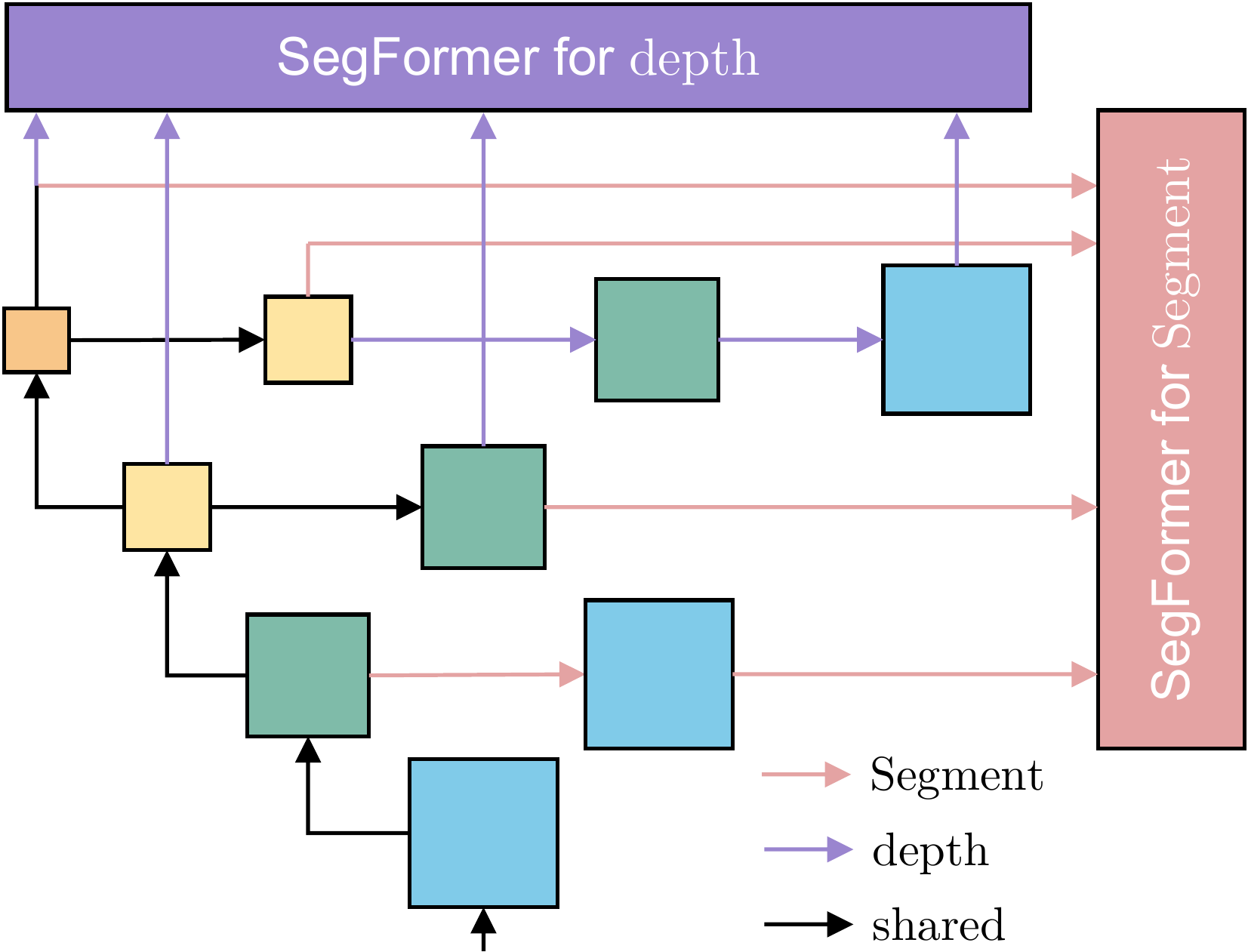} \\
\textbf{AutoTaskFormer with MLP heads (single-scale).} & \textbf{AutoTaskFormer with SegFormer \cite{xie2021segformer} heads (multi-scale)}
\end{tabular}
\end{center}
\caption{We visualized both the searched single-scale (Left) and multi-scale (Right) skeletons and their multi-task architectures for 2-tasks learning on \textbf{Cityscapes}. \label{fig:supp_city_archs}}
\end{figure*}

\begin{table*}[t]
\begin{center}
\small
\setlength{\tabcolsep}{3pt}
\aboverulesep=0.5ex
\belowrulesep=0.5ex
\begin{tabular}{c|ccc|ccccccccccccc}
\toprule
\multicolumn{1}{c|}{} & \multicolumn{3}{c|}{\textbf{Point predictions}}  & \multicolumn{12}{c}{\textbf{Dense predictions}} & \\
& \multicolumn{2}{c}{\footnotesize{Acc.$\uparrow$}} & \footnotesize{loss$\downarrow$} & \footnotesize{mIoU$\uparrow$} & \multicolumn{10}{c}{\footnotesize{MS-SSIM $\uparrow$}} & \multicolumn{2}{c}{\footnotesize{loss $\downarrow$}} \\
\cmidrule(lr){2-3} \cmidrule(lr){4-4} \cmidrule(lr){5-5} \cmidrule(lr){6-15} \cmidrule(lr){16-17} 
& \footnotesize{Obj.} & \footnotesize{Scene} & \footnotesize{Van. Pt} & \footnotesize{SegSem} & \footnotesize{$\mathcal{D}$e} & \footnotesize{$\mathcal{D}$z} & \footnotesize{\textit{E}2d} & \footnotesize{\textit{E}3d} & \footnotesize{$\mathcal{K}$2d} & \footnotesize{$\mathcal{K}$3d} & \footnotesize{$\mathcal{N}$} & \footnotesize{$\mathcal{PC}$} & \footnotesize{$\mathcal{R}$} & \footnotesize{\textit{AE}} & \footnotesize{Seg2d} & \footnotesize{Seg2.5d} \\
\midrule
Swin-T (single-task) & 47.9 & 60.3 & 1.567 & 31.0 & 86.6 & 86.6 & 94.5 & 73.0 & 96.6 & 56.4 & 75.7 & 67.3 & 68.7 & 96.0 & 0.446 & 0.150 \\
\midrule
task loss weight ($\lambda_k$) & 2 & 2 & 5 & 5 & 2 & 2 & 30 & 10 & 30 & 10 & 20 & 0.5 & 20 & 0.1 & 1 & 1 \\
\bottomrule
\end{tabular}
\end{center}
\caption{Results of single task baseline (Swin-T \cite{liu2021swin}) to train each task on Taskonomy. We provide the task weightings $\lambda_k$ for 16 tasks. \label{tab:sup_taskonomy}}
\end{table*}

\subsection{2 tasks learning on Cityscapes} Cityscapes \cite{cordts2016cityscapes} consists 5K high-resolution street-view images. We adopt 19-class annotation for semantic segmentation and estimate depth following \cite{liu2019end}. All images are resized to 256$\times$256 by random flipping and cropping during training, and are resized to 256$\times$512 during test. We train AutoTaskFormer as well as the other baseline vision transformers for 200 epochs. We use AdamW optimation with a base learning rate of 5e-4 for batch size of 8. We use cosine scheduler to decrease learning rate to 5e-6 and set the weight decay to 0.05 for regularization. The task loss weights $\lambda_k$ during training are set to 1 for both semantic segmentation and depth estimation.

We visualize both the single-scale and multi-scale skeletons and their multi-task architectures in Fig. \ref{fig:supp_city_archs}. It indicates AutoTaskFormer shares the certain stage of knowledge between tasks and owns the individual knowledge for a single task.

\subsection{3 tasks learning on NYUv2} NYUv2 \cite{silberman2012indoor} consists of 1449 images from 464 different indoor scenes. We follow \cite{liu2019end} to use 40-class semantic segmentation, depth estimation, and surface normal predictions in a 3 task scenario. We perform transfer learning on 3 tasks NYUv2 for 50 epochs. During training, we resize the input images to 224$\times$224 by randomly cropping and test on the 224$\times$224 center cropped images. We use AdamW optimation with a base learning rate of 5e-4 for batch size of 8 and cosine scheduler to decrease learning rate to 5e-6. The task weight losses $\lambda_k$ for semantic segmentation, depth estimation, surface normal predictions are set to 1, 3, 20 respectively.

\subsection{Detailed Cell Search Space\label{sec:cell}}

\begin{table}[t]
\begin{center}
\small
\setlength{\tabcolsep}{3pt}
\resizebox{0.47\textwidth}{!}{%
\renewcommand\arraystretch{1.05}
\begin{tabular}{l|c|cc|ccc}
\toprule
\#Param & Stride & Embed dim & Head Num & Depth & MLP ratio & Window \\
\midrule
\multirow{4}{*}{\makecell[l]{22.3 \\ $\sim$62.2M}} & 1/4 &  $\{$64, 96, 128$\}$ & $\{$2, 3, 4$\}$ & \multirow{4}{*}{$\{$2, 4$\}$} & \multirow{4}{*}{$\{$3.5, 4$\}$} & \multirow{4}{*}{$\{$5, 7, 9$\}$} \\
 & 1/8 & $\{$160, 192, 224$\}$  & $\{$5, 6, 7$\}$ & &  & \\
 & 1/16 & $\{$352, 384, 416$\}$  & $\{$11,12,13$\}$ & &  & \\
 & 1/32 & $\{$732, 768, 800$\}$ & $\{$23,24,25$\}$ & & & \\
 \midrule
\multirow{4}{*}{\makecell[l]{37.2 \\ $\sim$108.7M}} & 1/4 & $\{$96, 128, 160$\}$ & $\{$3, 4, 5$\}$ & \multirow{4}{*}{$\{$2, 4$\}$} & \multirow{4}{*}{$\{$3.5, 4$\}$} & \multirow{4}{*}{$\{$5, 7, 9$\}$} \\
 & 1/8 & $\{$192, 256, 320$\}$ & $\{$6, 8, 10$\}$ &  &  &  \\
 & 1/16 & $\{$448, 512, 576$\}$ & $\{$14, 16, 18$\}$ &  &  &  \\
 & 1/32 & $\{$960, 1024, 1088$\}$ & $\{$30, 32, 34$\}$ &  &  & \\
 \bottomrule
\end{tabular}
}
\end{center}
\caption{The cell search space of AutoTaskFormer. For example, for AutoTaskFormer of size 22.3$\sim$62.2M, we search embedding dimension from the set $\{$64,96,128$\}$. \textbf{Note}: The embedding dim and number of heads are set according to the stride of cells while the others are irrelevant to the stride. \label{tab:cell}}
\end{table}

According to the constraints on model parameters, we partition the cell space following the designs of Swin-S and Swin-B. The detailed design is presented in Tab. \ref{tab:cell}.

\subsection{Evolutionary Search \label{sec:evolution_search}}

We search for subnets $\alpha \in \mathcal{A}$ according to $\gamma(\alpha)$ that yield the best multi-task performance under different resource constraints via the evolution search. Specifically, we first pick 50 random subnets in the cell space as seeds and set the number of generations to 20. The top 10 subnets are picked as parents to generate the next generation by mutation and crossover. For a crossover, two randomly selected candidates are picked and crossed to generate a new one. For mutation, we first mutate depth and embedding dimension in each transformer layer with a probability of 0.4, and then mutate cell parameters in each block with a probability of 0.2 to produce new architecture.

\section{Supplementary Comparing Results}

\begin{figure*}[t]
\begin{center}
\setlength{\tabcolsep}{0pt}
\begin{tabular}{cccc}
\includegraphics[width=0.25\textwidth]{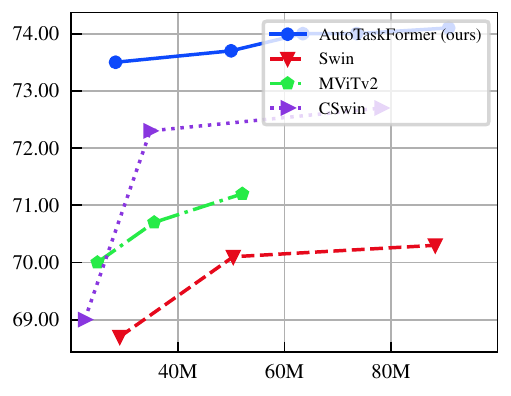} &
\includegraphics[width=0.25\textwidth]{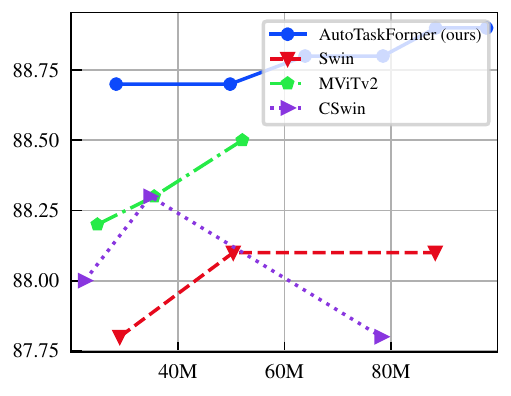} &
\includegraphics[width=0.25\textwidth]{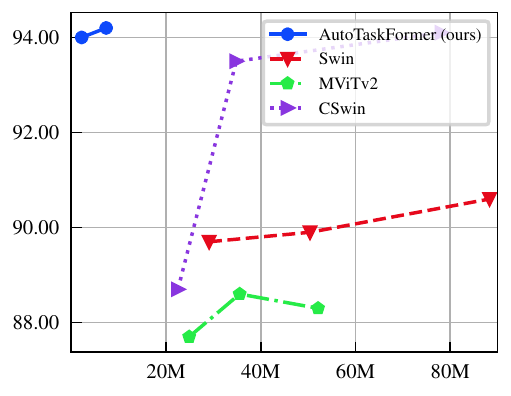} &
\includegraphics[width=0.25\textwidth]{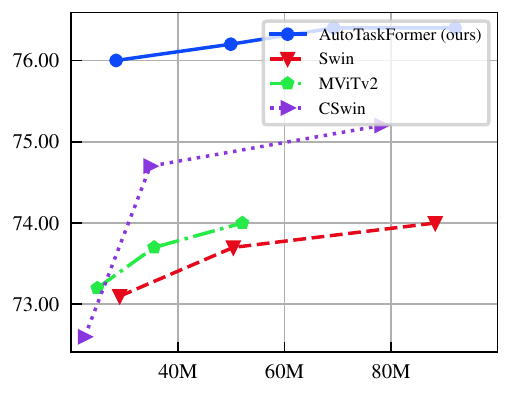} \\
\footnotesize{(a) Reshading (MS-SSIM)} & \footnotesize{(b) Depth Euclidean (MS-SSIM)} & \footnotesize{(c) Edge2d (MS-SSIM)} & \footnotesize{(d) Edge3d (MS-SSIM)} \\
\includegraphics[width=0.25\textwidth]{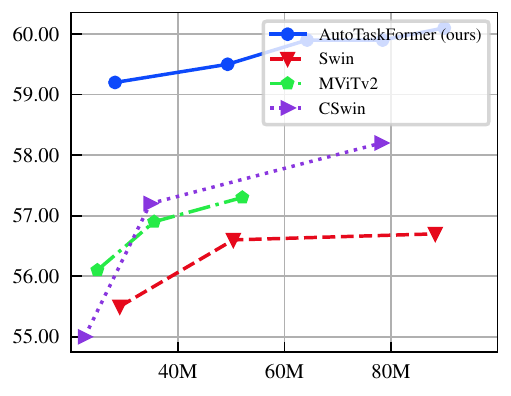} &
\includegraphics[width=0.25\textwidth]{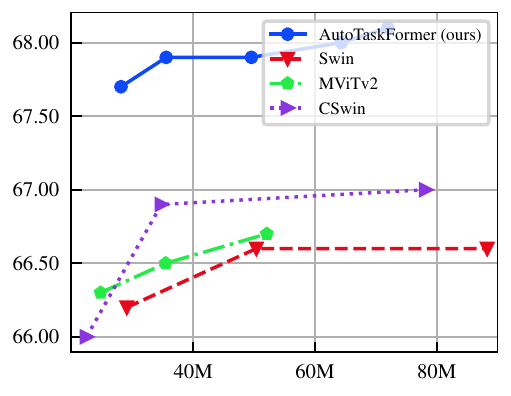} &
\includegraphics[width=0.25\textwidth]{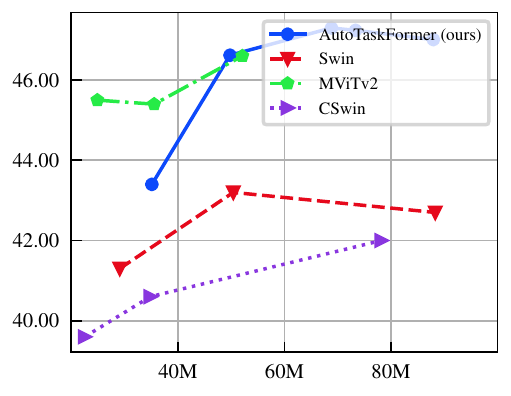} &
\includegraphics[width=0.25\textwidth]{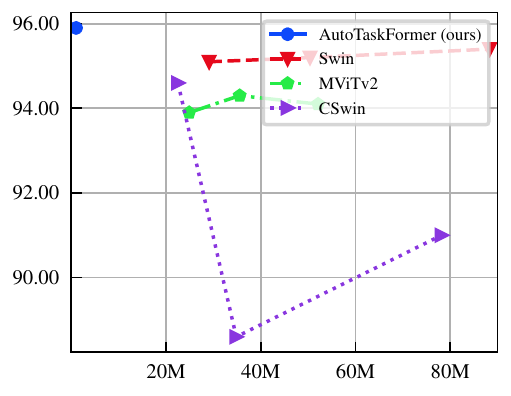} \\
\footnotesize{(e) Keypoints3d (MS-SSIM)} & \footnotesize{(f) Principal Curvature (MS-SSIM)} & \footnotesize{(g) Semanti Segmentation (mIoU)} & \footnotesize{(h) Auto Encoding (MS-SSIM)} \\
\includegraphics[width=0.25\textwidth]{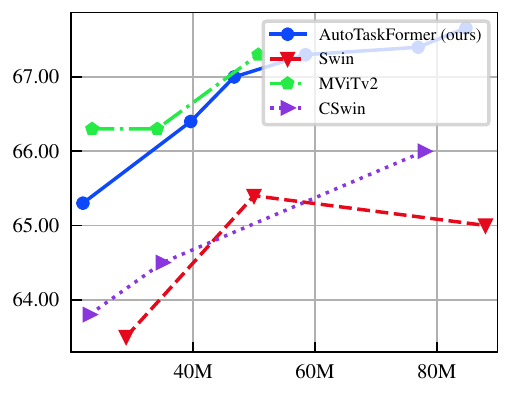} &
\includegraphics[width=0.25\textwidth]{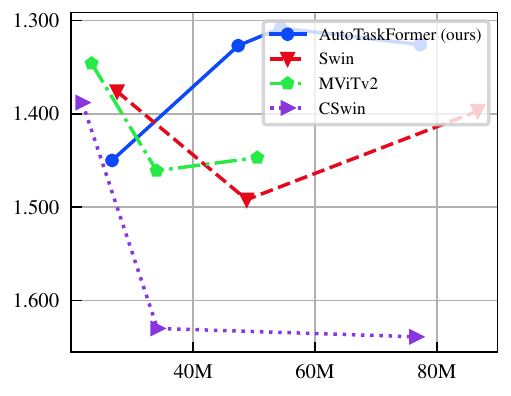} &
\includegraphics[width=0.25\textwidth]{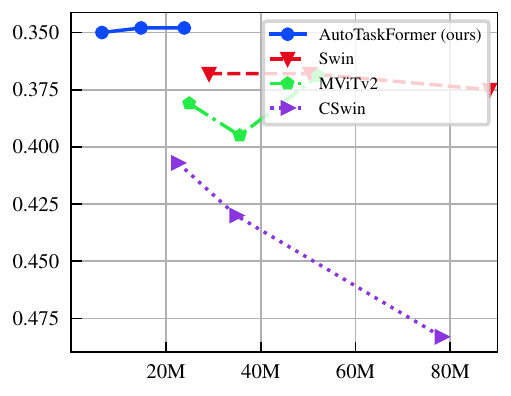} &
\includegraphics[width=0.25\textwidth]{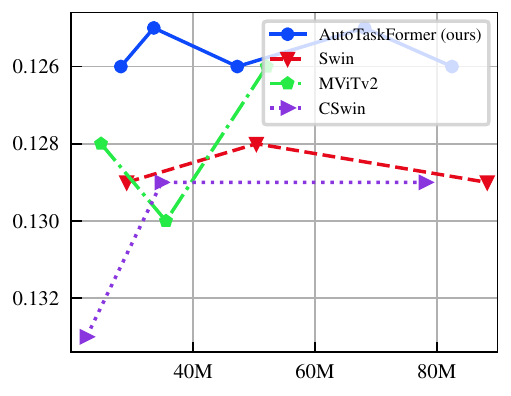} \\
\footnotesize{(i) Class Scene (Acc, \%)} & \footnotesize{(j) Vaninshing Point (loss)} & \footnotesize{(k) Unsupervised Seg2d (loss)} & \footnotesize{(l) Unsupervised Seg2.5d (loss)} \\
\includegraphics[width=0.25\textwidth]{figs/single_task/class_object.pdf} &
\includegraphics[width=0.25\textwidth]{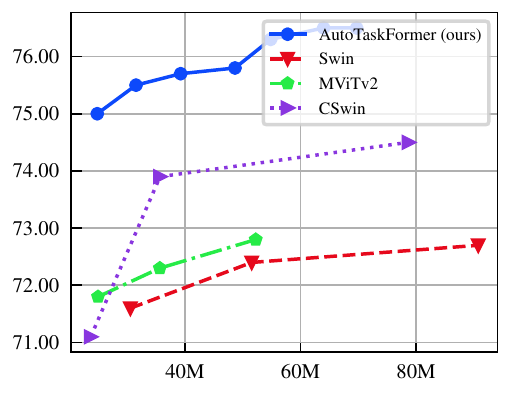} &
\includegraphics[width=0.25\textwidth]{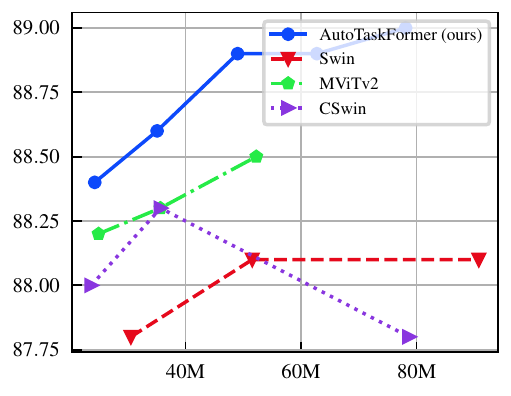} &
\includegraphics[width=0.25\textwidth]{figs/single_task/keypoints2d.pdf} \\
\footnotesize{(m) Class Object (Acc, \%)} & \footnotesize{(n) Surface Normal (MS-SSIM)} & \footnotesize{(o) Depth Zbuffer (MS-SSIM)} & \footnotesize{(p) Keypoints2d (MS-SSIM)} \\
\end{tabular}
\end{center}
\caption{The performances of single-task vision transformers after 16 tasks training on Taskonomy. We compare AutoTaskFormer with other handcrafted vision transformers on all 16 tasks. \label{fig:supp_task_pareto}}
\end{figure*}

We demonstrate that different tasks prefer different model complexities according to their favored skeletons. Thus, it is flexible to search smaller vision transformers for simple tasks. We supplement Fig. \ref{fig:task_pareto} with the other 14 tasks' performances in Fig. \ref{fig:supp_task_pareto} to show that once trained AutoTaskFormer can provide different vision transformers for various tasks under various resource constraints.

\section{Qualitative Results}

\begin{figure*}[t]
\begin{center}
\small
\newcommand{\centered}[1]{\begin{tabular}{l} #1 \end{tabular}}
\setlength{\tabcolsep}{3pt}
\begin{tabular}{m{2.5cm}m{14cm}}
\toprule
Swin \cite{liu2021swin}&
\includegraphics[width=0.8\textwidth]{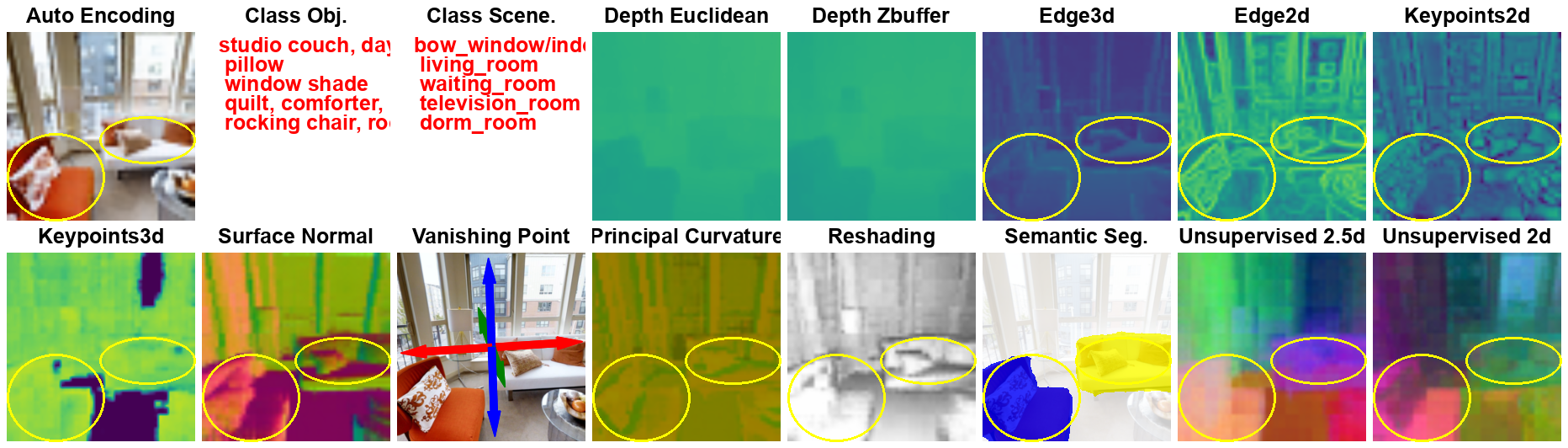} \\
\midrule
CSwin \cite{dong2022cswin} &
\includegraphics[width=0.8\textwidth]{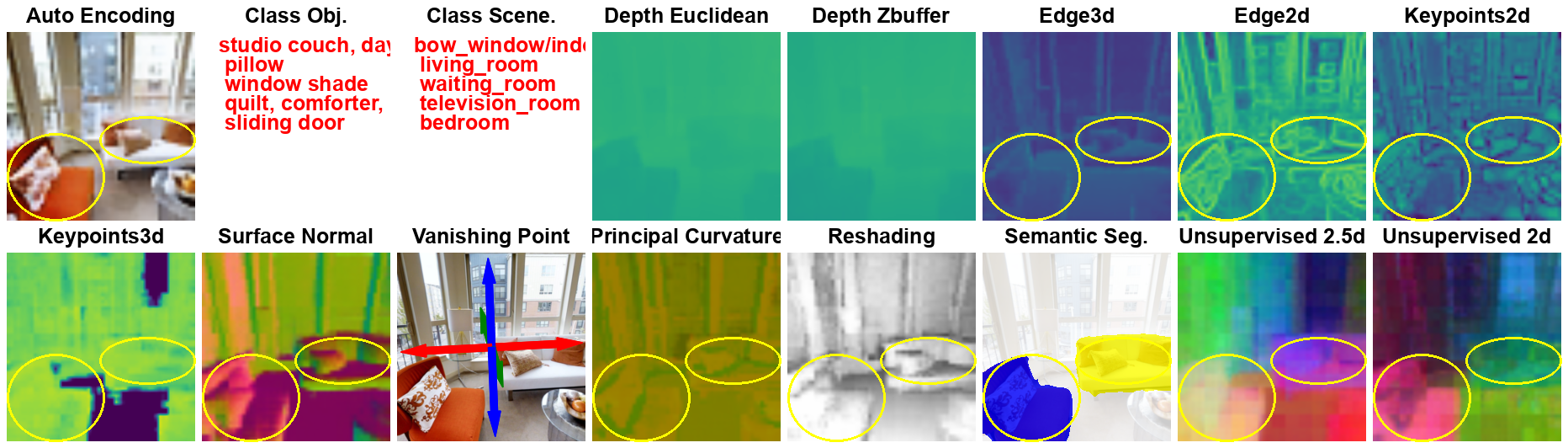} \\
\midrule
MViTv2 \cite{li2022mvitv2} &
\includegraphics[width=0.8\textwidth]{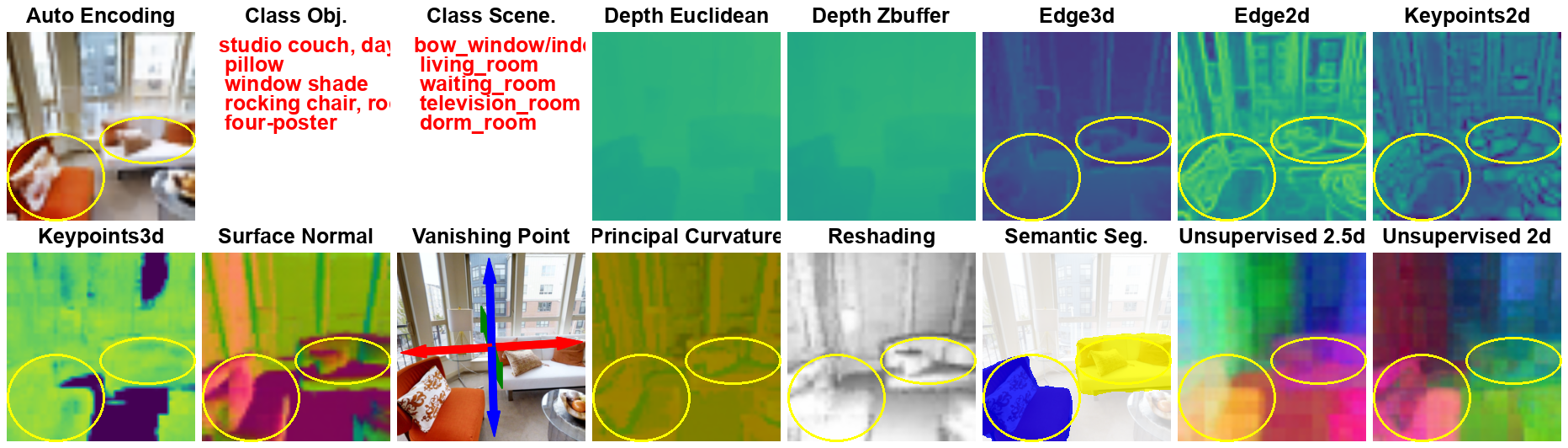} \\
\midrule
AutoTaskFormer (ours) &
\includegraphics[width=0.8\textwidth]{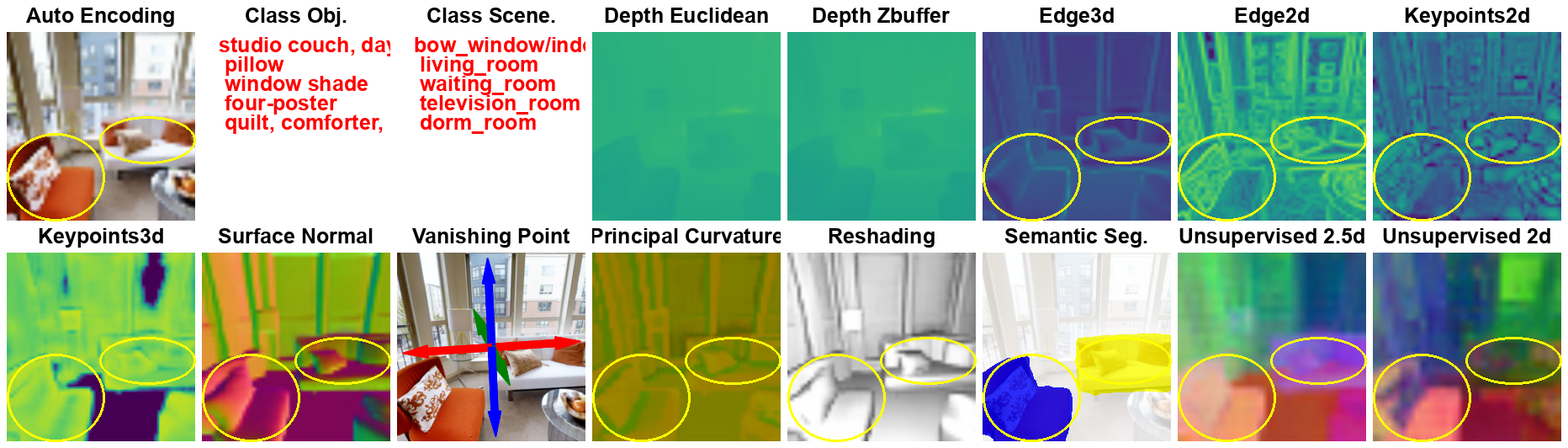} \\
\midrule
Ground Truth &
\includegraphics[width=0.8\textwidth]{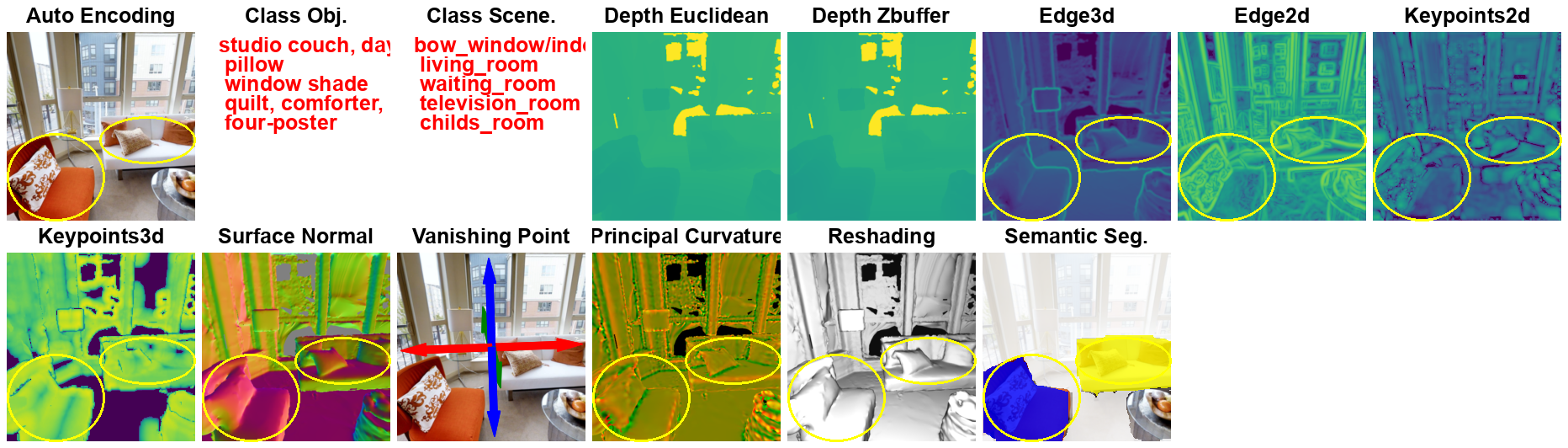} \\
\bottomrule
\end{tabular}
\end{center}
\caption{\textbf{Qualitative comparison for 16 tasks learning on Taskonomy ($\sim$40M parameters).} Our AutoTaskFormer outperforms other handcrafted multi-task vision transformers. Best seen on screen and zoomed within the yellow circled regions. \label{fig:supp_taskonomy}}
\end{figure*}

\begin{figure*}[t]
\begin{center}
\small
\newcommand{\centered}[1]{\begin{tabular}{l} #1 \end{tabular}}
\setlength{\tabcolsep}{3pt}
\begin{tabular}{m{2.5cm}m{14cm}}
\toprule
Swin \cite{liu2021swin}&
\includegraphics[width=0.8\textwidth]{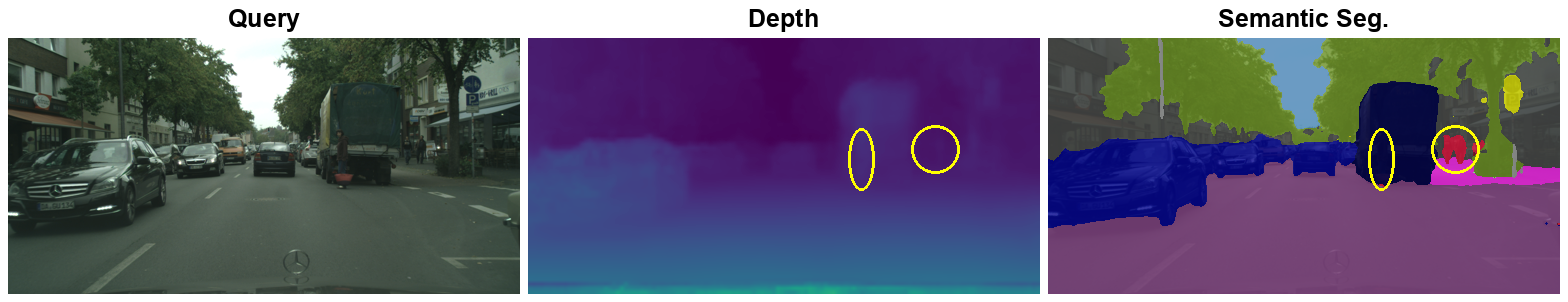} \\
\midrule
CSwin \cite{dong2022cswin} &
\includegraphics[width=0.8\textwidth]{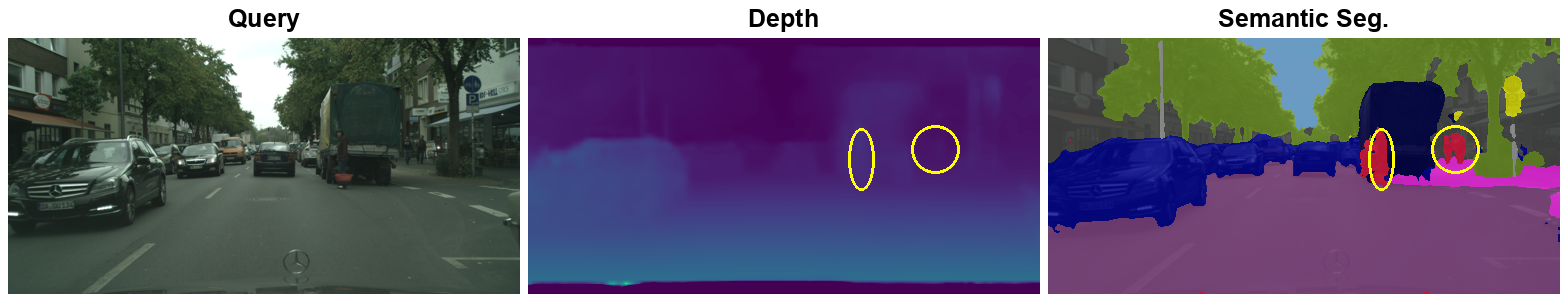} \\
\midrule
MViTv2 \cite{li2022mvitv2} &
\includegraphics[width=0.8\textwidth]{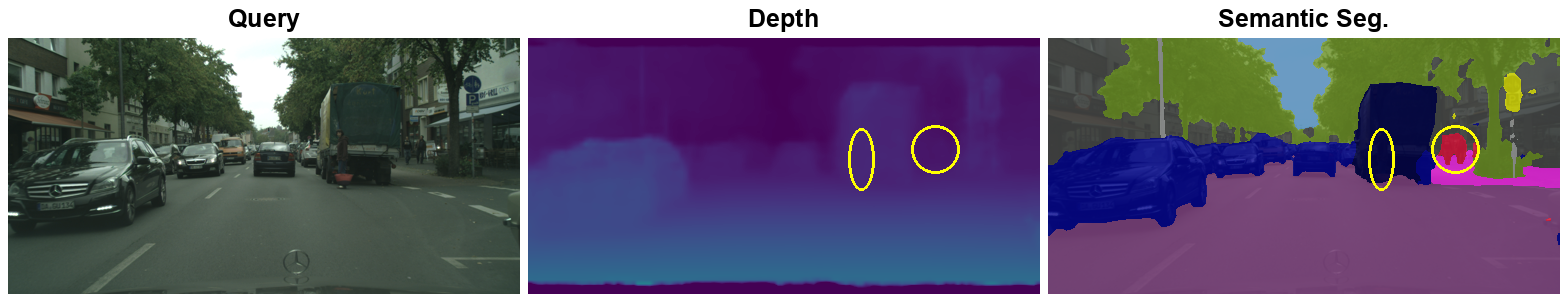} \\
\midrule
AutoTaskFormer (ours) &
\includegraphics[width=0.8\textwidth]{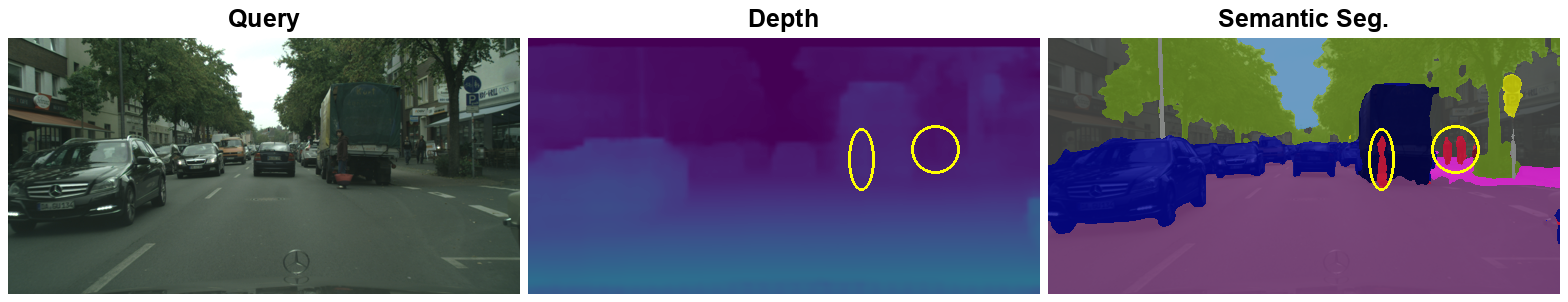} \\
\midrule
Ground Truth &
\includegraphics[width=0.8\textwidth]{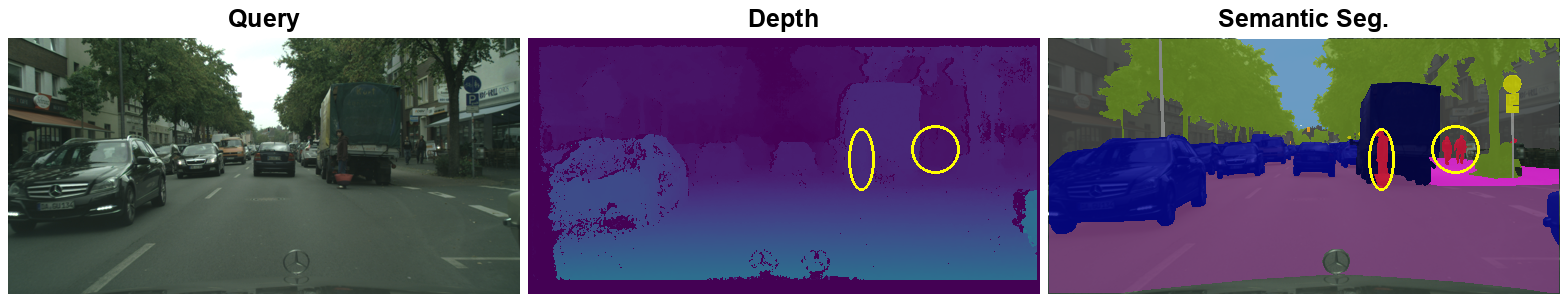} \\
\bottomrule
\end{tabular}
\end{center}
\caption{\textbf{Qualitative comparison for 2 tasks learning on Cityscapes ($\sim$30M parameters).} Our AutoTaskFormer outperforms other handcrafted multi-task vision transformers. Best seen on screen and zoomed within the yellow circled regions. \label{fig:supp_cityscapes}}
\end{figure*}

We qualitatively compare the results of our AutoTaskFormer with other handcrafted vision transformer baselines. Fig. \ref{fig:supp_taskonomy} shows the performance of 16 tasks learning on Taskonomy \cite{zamir2018taskonomy}. Fig. \ref{fig:supp_cityscapes} shows the results of 2 tasks learning on cityscapes \cite{cordts2016cityscapes} with SegFormer \cite{xie2021segformer} task heads.

\end{document}